\documentclass{article}
\usepackage{wrapfig}
\usepackage{makecell}
\usepackage{cite}
\usepackage{amsmath,amssymb,amsfonts}
\usepackage{algorithmic}
\usepackage{graphicx}
\usepackage{textcomp}
\usepackage{authblk}
\begin{document}
\title{A Clinical Evaluation of a Low-Cost Strain Gauge Respiration Belt and Machine Learning to Detect Sleep Apnea}

\author[1]{Stein Kristiansen (steikr@ifi.uio.no)}
\author[1]{\\Konstantinos Nikolaidis (konstan@ifi.uio.no)}
\author[1]{\\Thomas Plagemann (plageman@ifi.uio.no)}
\author[1]{\\Vera Goebel (goebel@ifi.uio.no)}
\author[2,3,4]{\\Gunn Marit Traaen (gtraaen@ous-hf.no)}
\author[5]{\\Britt {\O}verland (britt.overland@lds.no)}
\author[6,7]{\\Lars Aaker{\o}y (lars.aakeroy@stolav.no)}
\author[2,3,8]{\\Tove-Elizabeth Hunt (tovhun@ous-hf.no)}
\author[7,10]{\\Jan P{\aa}l Loennechen (jan.pal.loennechen@stolav.no)}
\author[6,7]{\\Sigurd Loe Steinshamn (sigurd.steinshamn@ntnu.no)}
\author[2]{\\Christina Holt Bendz (cholt@ous-hf.no)}
\author[2,8]{\\Ole-Gunnar Anfinsen (oanfinse@ous-hf.no)}
\author[2,3,4]{\\Lars Gullestad (lars.gullestad@medisin.uio.no)}
\author[3,9]{\\Harriet Akre (haakre@ous-hf.no)}

\affil[1]{Department of Informatics, University of Oslo, Oslo, Norway}
\affil[2]{Department of Cardiology, Oslo University Hospital, Rikshospitalet, Oslo, Norway}
\affil[3]{Institute of Clinical Medicine, Faculty of Medicine, University of Oslo, Oslo, Norway}
\affil[4]{KG Jebsen Center for Cardiac Research, University of Oslo, Oslo, Norway and Center for Heart Failure Research, Oslo University Hospital, Oslo, Norway}
\affil[5]{Department of Otorhinolaryngology, Head and Neck Surgery, Sleep Unit, Lovisenberg Diakonale Hospital, Oslo, Norway}
\affil[6]{Department of Thoracic Medicine, St. Olavs University Hospital, Trondheim, Norway}
\affil[7]{KG Jebsen Center of Exercise in Medicine, Department of Circulation and Medical Imaging, Faculty of Medicine and Health Science, Norwegian University of Science and Technology, Trondheim, Norway.}
\affil[8]{8 Department of Cardiology and Center for Cardiological Innovation, Oslo University Hospital, Rikshospitalet, Oslo, Norway}
\affil[9]{Department of Otorhinolaryngology, Head and Neck Surgery, Oslo University Hospital, Rikshospitalet, Oslo, Norway}
\affil[10]{Department of Cardiology, St.Olavs University Hospital, Trondheim, Norway}

\maketitle

\begin{abstract}
  Sleep apnea is a serious and severely under-diagnosed sleep-related respiration disorder characterized by repeated disrupted breathing events during sleep. It is diagnosed via polysomnography which is an expensive test conducted in a sleep lab requiring sleep experts to manually score the recorded data. Since the symptoms of sleep apnea are often ambiguous, it is difficult for a physician to decide whether to prescribe polysomnography. In this study, we investigate whether helpful information can be obtained by collecting and automatically analysing sleep data using a smartphone and an inexpensive strain gauge respiration belt. We evaluate how accurately we can detect sleep apnea with wide variety of machine learning techniques with data from a clinical study with 49 overnight sleep recordings. With less than one hour of training, we can distinguish between normal and apneic minutes with an accuracy, sensitivity, and specificity of 0.7609, 0.7833, and 0.7217, respectively. These results can be achieved even if we train only on high-quality data from an entirely separate, clinically certified sensor, which has the potential to substantially reduce the cost of data collection. Data from a complete night can be analyzed in about one second on a smartphone.
\end{abstract}


\section{Introduction}
\label{sec:introduction}
Sleep Apnea (SA) is a common and strongly under-diagnosed sleep-related respiratory disorder with severe health implications. It affects the natural breathing cycle during sleep and the disruption of normal airflow causes a decrease of oxygen in the blood. If the oxygen level is too low, the brain will force an awakening in order to resume normal breathing. The patient will most likely not remember the continuous awakenings. With such disrupted sleep, the person will in some cases never go into deep sleep, resulting in daytime sleepiness and fatigue. When left untreated, SA can lead to serious medical conditions including cardiovascular diseases such as stroke, metabolic syndromes such as diabetes, and psychological disorders such as depression and anxiety \cite{young2004risk, punjabi2008epidemiology, huang2008clinical}. SA patients are also more prone to accidents with for instance motor vehicles \cite{teran1999association}.

One factor that contributes to SA being severely under-diagnosed is that feeling tired during the day is normal for many people and the patients typically have no recollection of the nightly awakenings. Therefore, patients can typically only describe vague symptoms to the physician that do not necessarily indicate the need for a clinical diagnosis with polysomnography (PSG) in a sleep laboratory (the gold standard for SA diagnosis) or polygraphy (PG) via unattended sleep monitoring at home. A low-cost solution enabling everyone to perform sleep monitoring and analysis independently at home could motivate patients to visit a physician before serious health implications are manifested and provide a better data foundation to physicians for decision making.

In our earlier research we found that machine learning (ML) can be used to detect SA events in sleep monitoring data. The performance is rather high for PSG and PG data, even if only a single physiological signal is used. For example, using data from a single respiration belt on the abdomen we could achieve an accuracy, sensitivity, and specificity of 0.9330, 0.9570, and 0.9160 with PSG data \cite{kristiansen2018data} and 0.8349, 0.8244, and 0.8452 with PG data \cite{kristiansen2020comparing}. Furthermore, we could demonstrate through controlled laboratory experiments that some low-cost respiration belts (based on strain gauge technology) achieve a data quality that comes close to the data quality of the respiration belts that are used for clinical diagnosis. The latter type of respiration belts is much more sophisticated and typically based on respiratory inductance plethysmography (RIP). For example, we achieved with the low-cost sensor from Sweetzpot, called Flow, a positive predictive value for breath detection accuracy of 99.16\%, using the RIP belt from Nox T3 as reference in the lab experiments \cite{loberg2018quantifying}.

In fall 2018, we had the unique opportunity to test the Flow sensor with real patients that underwent each two nights of PG with the Nox T3. This was performed as part of a large clinical study, called A3 study, at the Oslo University Hospital and St. Olavs Hospital, Trondheim \cite{traaen2020prevalence}. A total of 579 patients were involved in the A3 study and the last 29 patients used the Flow sensor and the Nox T3 for sleep monitoring at the same time. We collected recordings from at least two nights from 15 of the patients, resulting in recordings from a total of 49 nights. A sleep expert scored the Nox T3 data by labelling time periods that contain apneic events. We use these labels for the Flow data after synchronizing the Nox T3 and the Flow recordings. The Nox T3 and Flow data sets are used in this paper to investigate two basic research questions (RQ): (1) How accurately can ML models estimate SA severity with data from the Flow sensor?, and (2) Is the resources required for data analysis with an ML model sufficiently low to be performed on a smartphone?

This is to the best of our knowledge the first time a strain gauge breathing sensor is used in a clinical study with data from unattended home monitoring, and where an in-depth study of SA detection performance of ML with such data is presented. Several important results and insights are gained in this study. First, despite several issues with the Flow sensor, we achieve an accuracy, sensitivity, and specificity of 0.7609, 0.7833, and 0.7217 of classifying minutes as normal or apneic, with less than an hour of training. In one setting, when the ML-classifier predicts that an individual has moderate SA (more than 15 disrupted breathing events per hour), the probability of this actually being the case doubles relative to the prior probability. A positive test outcome is therefore useful for screening purposes. Second, we find that the resource consumption of data analysis is no issue for today's smartphones. Third, mixed training, i.e., using Flow data together with high-quality Nox-T3 data to train a model for Flow data classification, positively impacts the achievable classification performance; and using only Nox-T3 data as training data leads to comparatively good performance. This implies that to train a good classifier for another strain gauge sensor might not require collecting a large amount of sleep data with that sensor.

The remainder of this paper is structured as follows: In Section II we present the background and in Section III the data acquisition and pre-processing. In Sections IV and VII we describe the experiments and results. Section VIII concludes this paper with a discussion of the results and impact.

\section{Background}
\label{sec:background}
SA is a common sleep-related respiration disorder that affects the natural breathing cycle during sleep with periods of reduced respiration (called hypopneas) or no airflow at all (called apneas). The most common type is Obstructive SA (OSA), where the airway is physically obstructed, resulting in a reduction or a complete stop of air passing through. The less common Central SA (CSA) is caused by the lack of proper breathing signals from the brain. Disruptions of normal airflow causes a decrease of oxygen in the blood, commonly referred to as the oxygen de-saturation. When untreated, SA can lead to serious health implications. Hypertension, stroke and other cardiovascular diseases, metabolic syndromes such as diabetes, depression and anxiety have all been linked to SA \cite{young2004risk, punjabi2008epidemiology, huang2008clinical}. With the daily drowsiness, SA patients are more prone to accidents, especially with motor vehicles \cite{teran1999association} and other activities that require a high level of concentration.

An estimated 936 million individuals aged 30-69 years (men and women) are estimated to suffer from OSA and 426 millions of these cases are categorized as moderate or severe \cite{benjafield2019estimation}. In Norway, 16\% of the 30-65 aged population suffer from SA and 8\% are estimated to have moderate or severe SA\cite{hrubos2011norwegian}. Unfortunately, SA is often diagnosed very late or not at all. It is estimated that 70-80\% of those affected remain undiagnosed \cite{punjabi2008epidemiology}. One factor that contributes to SA being severely under-diagnosed is that feeling tired during the day is normal for many people and the patients have typically no recollection of the nightly awakenings. Patients can typically only describe vague symptoms to the physician that do not necessarily indicate the need to perform PSG.

\subsection{SA Diagnosis}
The gold standard for diagnosing SA is via PSG in a sleep laboratory \cite{punjabi2008epidemiology}. It requires the patient to stay overnight and record various physiological signals during sleep, such as the electrocardiogram, electroencephalogram, electromyogram, electrooculogram, oxygen saturation, heart rate, blood pressure and respiration from the abdomen, chest, and nose. A sleep expert is required to carefully analyze the data according to strict guidelines like those provided by the American Academy of Sleep Medicine (AASM) \cite{berry2012aasm}. This process involves identifying and denoting the beginning, end, and type of all periods with disrupted breathing. The two types of disrupted breathing are called apneas and hypopneas. Apneas are defined as a time period with a drop of airflow of at least 90\% lasting for at least 10 seconds. Hypopneas are defined as a time period of at least 10 seconds with a drop of airflow of at least 30\% breathing followed by a drop of oxygen saturation of at least 3\%. A diagnosis is given based on the average number of such periods per hour, called the Apnea-Hypopnea Index (AHI). The AHI is used to classify the patient into one of four classes: (1) normal, no SA: AHI $<$ 5, (2) mild SA: 5 $\leq$ AHI $<$ 15, (3) moderate SA: 15 $\leq$ AHI $<$ 30, and (4) severe SA: AHI $\geq$ 30.

PSG is very expensive, time consuming, uncomfortable, and few sleep labs exist. A sleep expert is required to analyze and score the data. Type III and IV portable monitoring systems are intended to alleviate these problems and allow unattended sleep monitoring at home. There exists a wide range of PG devices that are certified for clinical SA diagnosis \cite{mendoncca2018devices,collop2011obstructive}. With PG many more people can be analyzed. PG is ideal for preliminary screening purposes to determine whether a PSG should be carried out.

Respiration belts are used in PSG and PG to measure respiratory effort from the abdomen and chest. Respiratory Inductance Plethysmography (RIP) is the gold standard \cite{berry2012aasm}, where coils are sewn into the flexible belts in a sinusoidal pattern and connected to an oscillator. Changes in the lengths of the belts during respiration alters the self-inductance of the coils which results in changes in the oscillations \cite{cohn1982respiratory}. When properly calibrated, these changes accurately reflect the tidal volume of breathing \cite{chadha1982validation}. The Flow belt used in our study is based on the much less expensive strain gauge technology \cite{sweetzpot2020}. Variations in the lengths of the belts during breathing are measured as changes in the tension of the belt at a single point, i.e., as changes in the electrical resistance in the strain gauge at the front of the chest or abdomen. While RIP belts measure actual changes in the belt length, strain gauges only estimate such changes at a single-point and are as such susceptible to errors due to local variations in tension. Unlike the RIP technology, strain gauge has not been subject to rigorous clinical validation for sleep monitoring. 

\subsection{Automatic Analysis with ML}
\label{section:autoanalysis}
The achievable performance of automatic analysis of sleep data from clinically validated devices is well-known. There is a large body of research on automatic SA detection \cite{alvarez2015computer, faust2016review, pombo2017classification, uddin2018classification, mendonca2018review, mostafa2019systematic}. Many devices are intended for home monitoring \cite{collop2011obstructive}, and most have support for (semi-)automatic analysis \cite{mendoncca2018devices}. They are however much more expensive than the consumer electronics used in this work, and normally use proprietary, custom-tailored analysis solutions that cannot be leveraged for the large and growing variety of sensors in the consumer market. Among the papers that use ML for data analysis, almost all use data from PSG, commonly obtained from one of 23 online databases \cite{mostafa2019systematic}. We are only aware of three works that use data from unattended home monitoring \cite{alvarez2016automated, gutierrez2018evaluation, alvarez2020machine}, but these use expensive, clinically validated devices and other signals than in our work.

Accuracy, sensitivity, and specificity are commonly used to quantify the accuracy of classifying subjects into different SA severity groups (called \emph{subject-based}) or of classifying fixed-size periods of the data as normal or apneic (called \emph{period-based}). Period-based classification can be used to estimate SA severity by counting the number of apneic periods per hour. A commonly used period size in related works is 60 seconds.

Most of the above mentioned studies use several sensors at once. Using one or two sensors is also relatively well studied for the signals oxygen saturation \cite{al2014classifying, kaimakamis2009screening, garde2015pulse,hafezi2019sleep} and ECG \cite{kaimakamis2009screening}. Recently, accelerometers \cite{al2014classifying,camci2017sleep,ferrer2019automatic,hafezi2019sleep}, microphones \cite{al2014classifying, nandakumar2015contactless,camci2017sleep,castillo2019entropy,hafezi2019sleep}, radars \cite{javaid2015towards}, and pressure sensors \cite{zhang2013real} have been extensively studied. The use of respiratory belts alone has not been equally well studied, especially with only one belt. We are only aware of six such works. In two of these, disrupted breathing is simulated while awake \cite{nepal2002apnea,dehkordi2012monitoring}, which yields only approximate results. The remaining four use data from sleeping patients in a controlled PSG setting \cite{elmoaqet2020deep,van2020portable,lin2016sleep,tsouti2020development}, which is not ideal to study the performance for unattended home monitoring. Two of these use RIP belts \cite{elmoaqet2020deep, tsouti2020development}, and two are based on less expensive piezo and bioimpedance technologies and are thereby more similar to our work \cite{van2020portable, lin2016sleep}. A period-based accuracy of 72.8\% and 73.8\% was achieved with bioimpedance \cite{van2020portable} (25 subjects) and piezo sensors \cite{lin2016sleep} (34 subjects) at the abdomen, respectively. A period-based accuracy of 84.4\% was obtained with a single RIP belt on the abdomen (17 subjects) \cite{elmoaqet2020deep}, and a subject-based accuracy 74\% was achieved (12 subjects) \cite{tsouti2020development}.

Our work contributes to existing knowledge by using data from a clinical study in an unattended home monitoring setting to evaluate the ML performance with data from low-cost respiration belts. In particular, it provides insight into the performance of using strain-gauge technology for this purpose. By conducting a systematic, comparative analysis comprising a wide variety of modern deep ML techniques, we identify the best performing and least resource demanding ML-technique. We also evaluate the feasibility of running this model on a smartphone. We contribute with insights into the performance improvements obtainable by mixing of high- and low-quality data during the training phase, and on the trade-offs between using balanced versus imbalanced training sets for subject-based classification.

\section{Data Acquisition, Pre-Processing, and Method}
This section describes the data acquisition and the pre-processing to (1) handle data quality issues and (2) prepare the data for the ML studies, and the evaluation method.

\subsection{Data Acquisition}
\label{section:dataaquisition}
The A3 study is a large clinical study performed at the Oslo University Hospital and St. Olav Hospital, Trondheim which investigates the prevalence of SA among patients with paroxysmal atrial fibrillation. In this study, 579 subjects with paroxysmal atrial fibrillation were recruited to perform unattended sleep monitoring at home (most of them for two nights) with the Type III monitor Nox T3 \cite{website:nox-t3}. Each individual was given brief training on sensor placement and use of the Nox-T3 device before bringing the device home to undergo unattended sleep monitoring. We got the unique opportunity to additionally collect with the Flow sensor the respiration data from the 29 patients. These patients were instructed on how to attach the Flow sensor side-by-side to the abdomen RIP belt of the Nox-T3 and how to use the SleepRecorder app on an Android phone for data acquisition. The patient then performed the sleep monitoring at home with both monitoring solutions simultaneously. The data from Nox-T3 was scored by a sleep expert and used for the A3 study. Both the Nox-T3 and Flow data was exported as CSV files and a peak detection algorithm was used to synchronize their timelines \cite{loberg2018quantifying}. This way, the scoring from the sleep expert could be used to label periods of Flow data as either apneic or normal. Among the 29 individuals from which we collected Flow data, 21 are male and 8 are female. The mean age and BMI was 61.1 (SD 10.9) and 28.2 (SD 4.8), respectively. The number in the SA severity classes mild, moderate, and severe is 13, 10, and one, respectively. Five individuals had an AHI below five, and thus belong to the class of normal individuals (no SA).

\subsection{Pre-Processing}
\label{section:preprocessing}
Movement during the night can result in short periods with motion artifacts in the sleep data which make the corresponding data useless for SA detection. This may subsequently result in a permanent change in the tension in the belt causing a sudden baseline shift in the recorded signal. The sensing technology in RIP belts is superior to strain gauge belts for such situations. The Flow sensor in addition suffers from gradual baseline wandering and drift. Nox Medical has spent a substantial effort to eliminate all baseline issues in the Nox-T3 data. We design a simple Baseline Adjustment (BLA) procedure to reduce these issues for the Flow data where we standardize the Flow data using a one-minute sliding window over the entire recording \cite{hamborg2020}.

\begin{table}
  \centering
  \begin{tabular}{ c | c | c || c | c | c}
    \textbf{Exp.}&\textbf{Training set}&\textbf{Test set}&\textbf{Exp.}&\textbf{Training set}&\textbf{Test set}\\
    \hline
    1.1&F&F&2.3&NF, FB&FB\\
    1.2&FB&FB&2.4&N, FB&FB\\
    1.3&NF&NF&3&FB&FB\\
    2.1&N&FB&4&FB&FB\\
    2.2&N, FB&FB&&&\\
  \end{tabular}
  \caption{Data sets used in the experiments.}
  \label{table:datasets}
\end{table}

We refer to data from the Flow sensor as \emph{Flow data}, and Flow data with and without BLA treatment as $FB$ and $F$, respectively. Each of these contains 235 hours of data. We also use two sub-sets of the data from the Nox-T3 (\emph{Nox data}). The first ($N$) contains the entire A3 data set from 579 individuals. Since some recordings contain many artifacts, we remove all recordings with more than 20\% artifacts, leaving a total of 7408.85 hours of data. The second sub-set, called $NF$, contains only the portion of the Nox data that was obtained at the same time as the Flow data (and was synchronized with it). This data set is of the same size as $FB$. Table \ref{table:datasets} shows the data sets used for training and testing in our experiments. How they are used is described in the experiment descriptions below. The sampling rate of Flow and Nox-T3 sensors is 10Hz\footnote{In practice, this sampling rate may vary slightly over time \cite{hamborg2020}.} and 100Hz, respectively. Since breathing movements span several seconds, we can down-sample the data without loss of important respiratory changes. We down-sample the data to 1Hz using the mean value of each second, and standardize the resulting sample values.



\subsection{Evaluation Method}
\label{section:experiments}
The experiments in this work are designed to address the two overall research questions (RQ) mentioned in the introduction. Experiments 1, 2, and 3 address RQ1 (ML performance), and Experiment 4 addresses RQ2 (feasibility of using a smartphone for inference). Experiments 1 and 2 study performance in accordance with standard ML methodology for comparability with related works. Experiments 3 and 4 focuses on the performance in our application domain. We describe the aspects common to all experiments in this section, and experiment-specific details in Sections \ref{section:exp1} to \ref{section:exp4}. The experiments are conducted on a computer with a 56-core, 2.2Ghz, Intel Xeon Gold 5120 CPU, 128 GB RAM, and four Nvidia GTX 2080 Ti GPUs. In Experiment 4, we also use a Samsung Galaxy Tab S3.

All experiments except Experiment 3 (see Section \ref{section:exp3}) are conducted using 10-fold cross-validation (CV). It is common to randomly shuffle the data set before CV (called \emph{pre-shuffling}). This is not done in our experiments, since it may result in unrealistically high classification performance in our envisioned scenario. We assume that a pre-trained classifier is used to estimate the SA severity of an individual from which data was not available during training. With pre-shuffling, training data will contain a significant amount of samples from the same individuals as those in the test set.

We estimate SA severity via binary classification of periods. For comparability with related works, we classify each 60-second period as either \emph{apneic} or \emph{normal}\footnote{We do not distinguish between obstructive, central, or mixed apneas and hypopneas in this work, as it does not affect the AHI, which is the main metric to score SA severity.}. Each data set with Flow data ($F$, $FB$, and $NF$) has 2988 apneic periods and 10117 normal periods, while $N$ has 114009 apneic and 330522 normal periods. Due to the imbalanced class distributions, we balance the data set using majority sub-sampling before dividing it into 10 folds, if not explicitly stated otherwise. We discard the final periods to ensure equal-sized folds.

Training is performed via mini-batch learning with a batch size of 1000 periods. The learning rate fixed at 0.001, as no improvement was experienced with other learning rates during prior testing. 30\% of the training set is used as a validation set. We train all classifiers for 500 epochs as this was sufficient for all architectures to maximize performance during prior testing. For all Neural Networks (NN), we use the cross-entropy loss function and the Adaptive Moment Estimation (ADAM) optimizer.

We quantify classification performance using Cohens $\kappa$ metric \cite{cohen1960}. It is ideal for comparing classifiers as it accurately summarizes most aspects of prediction performance and accounts for random chance \cite{cohen1960}. In addition, we present accuracy, sensitivity, specificity since these have more intuitive semantics that can help understand the clinical relevance of the results. The presented results are those from the epoch with the highest $\kappa$ on the validation set. 

\section{The Impact of ML Technique and BLA}
\label{section:exp1}
In Experiment 1, we study the classification performance with the Flow sensor and state-of-the-art ML, and how this performance differs with ML-technique and BLA. We show that we can obtain an accuracy of 0.7609 with BLA. The performance is limited by the relatively simple and inexpensive construction of the Flow sensor and the simplicity of BLA. A key finding is that classifiers based on Convolutional NN (CNN) outperform the others in several respects. First, they rank among the best-performing classifiers in terms of classification performance - this is the architecture achieving the above mentioned accuracy. Second, they appear to be the most resilient to baseline shift and wandering. Third, they require an order of magnitude less computational resources than recurrent NN. We also find that BLA leads to substantial performance improvements for nearly all classifiers.

These findings are obtained by two sub-experiments: one without BLA using $F$ (Experiment 1.1) and one with BLA using $FB$ (Experiment 1.2). As a benchmark, we perform one experiment with $NF$, i.e., with only high-quality Nox data (Experiment 1.3). In each experiment, the different folds from the same data set is used for training and test via 10-fold CV. We include a wide range of ML classifiers that differ in architecture, approach, size, and sophistication. Most are based on modern deep NN. For comparison, we also include two traditional techniques that are commonly used in related works, i.e., Random Forest (RF) and Multi-Layer Perceptrons (MLP). The MLP are feed-forward NN with one hidden layer, and all layers densely connected. The deep NN architectures include CNN, Long Short-Term Memory (LSTM), and Gated Recurrent Unit (GRU). Since we use temporal data, for which LSTM are designed, we include several improvements and modifications of LSTM. These include BI-directional LSTM (BILSTM), Stacked BILSTM (SBILSTM), and SBILSTM With Attention (BIWALSTM) \cite{wang2016attention}. All classifiers are instantiated in the three sizes small, medium, and large. Classifiers are referenced with the acronyms of their architecture with a post-fix -S, -M, and -L for small, medium, and large sized classifiers, respectively. The parameter values for each classifier are presented in Table \ref{table:classifiers}. Some parameters are fixed because other values did not lead to performance improvements during prior testing and/or because the chosen value is commonly used in related works. For the same reasons, we use dropout only in dense layers in MLP and CNN, and no batch normalization.

\begin{table}
    \begin{tabular}{ l | l | l | l | l}
    \textbf{Classifier}&\textbf{Hyperparameter}&\textbf{Small}&\textbf{Medium}&\textbf{Large}\\
    \hline
    MLP&Hidden nodes&50&100&200\\
    \emph{Hidden layers: 1}&&&&\\
    \hline
    CNN&Conv. layers&3&4&5\\
    \emph{Kernel size: 5}&Filters/conv. layer&32-128&32-256&32-512\\
    &Dense layers x nodes&1x128&1x256&2x256\\ 
    \hline
    LSTM&Hidden state size&50&100&200\\
    \hline
    GRU&Hidden state size&50&100&200\\
    \hline
    BILSTM&Hidden state size&50&100&200\\
    \hline
    BIWALSTM&Hidden state size&50&100&200\\
    \hline
    SBILSTM&Layers&1&3&5\\
    \emph{Hidden state: 100}&&&&\\
    \hline
    RF&Trees&50&200&500\\
    \emph{Max. nodes/tree: 100}&&&&\\
    \hline
    \end{tabular}
\caption{Classifier parametrization.}
\label{table:classifiers}
\end{table}

\begin{figure}
  \centering
  \includegraphics[width=0.75\textwidth]{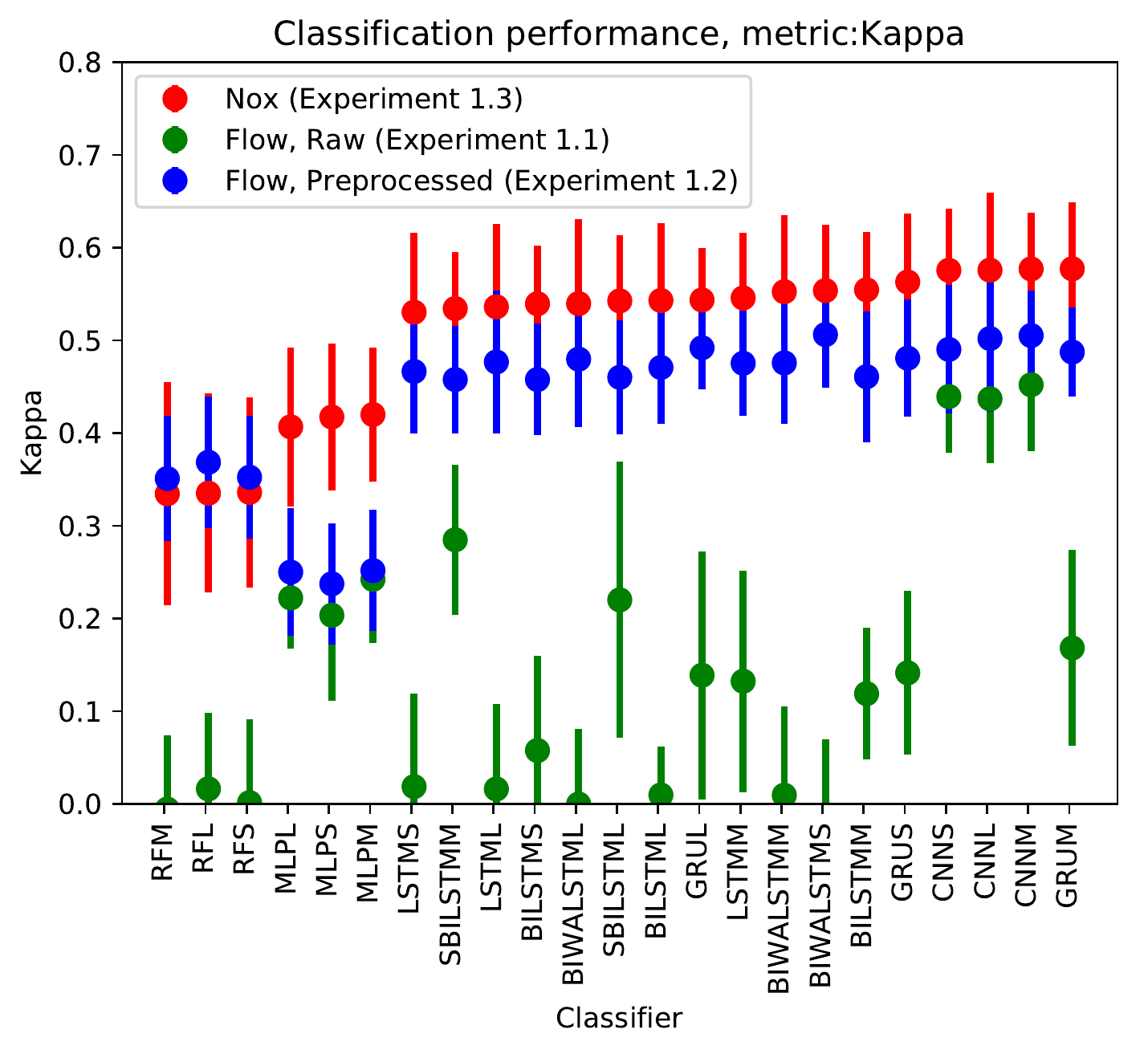}
  \caption{$\kappa$ for Experiment 1.}
  \label{figure:exp1}
\end{figure}

\begin{figure*}
  \centering
  \begin{tabular}{ccc}
     \makecell{
       \includegraphics[width=0.30\textwidth]{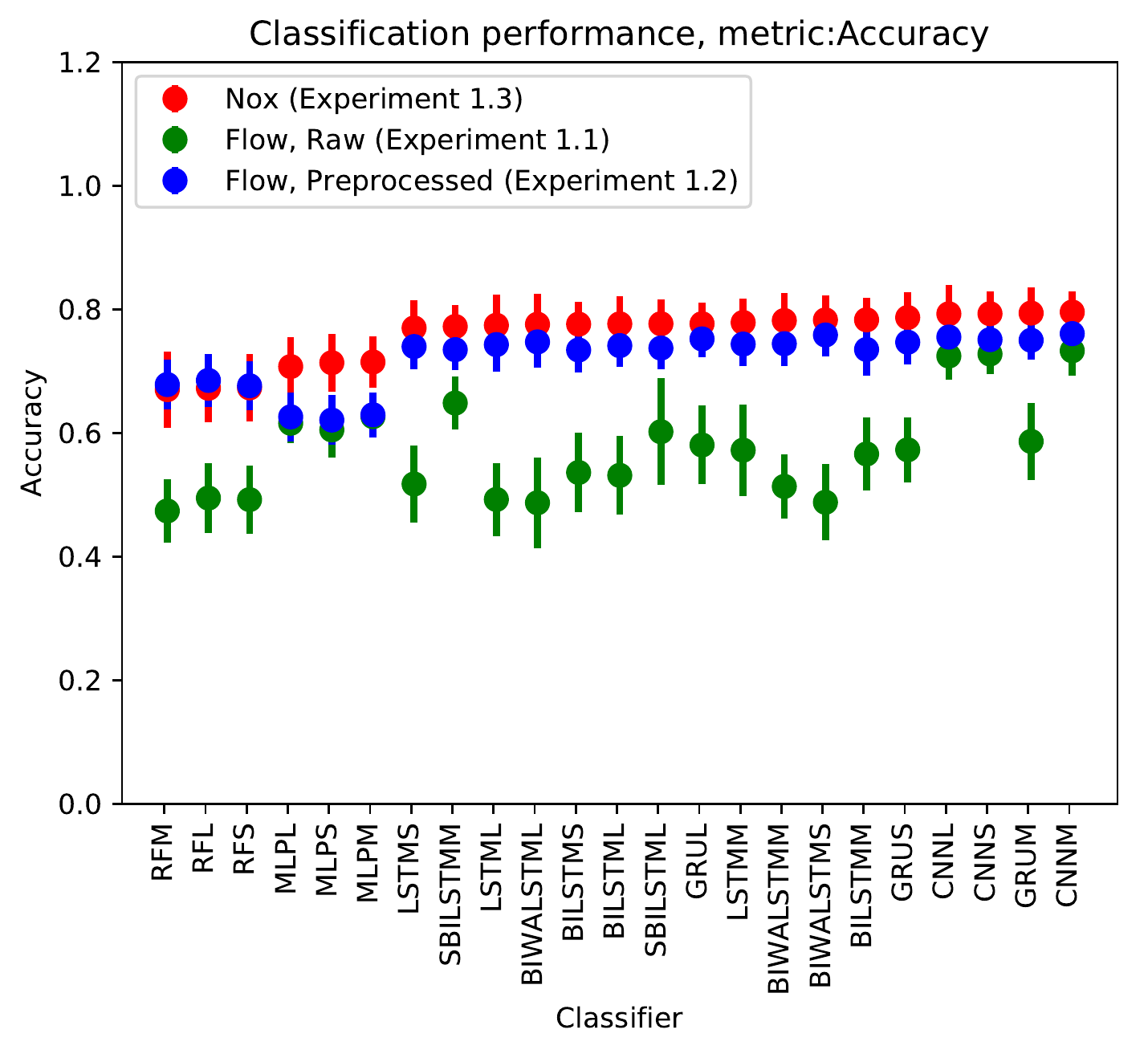}\\(a)}&
     \makecell{
       \includegraphics[width=0.30\textwidth]{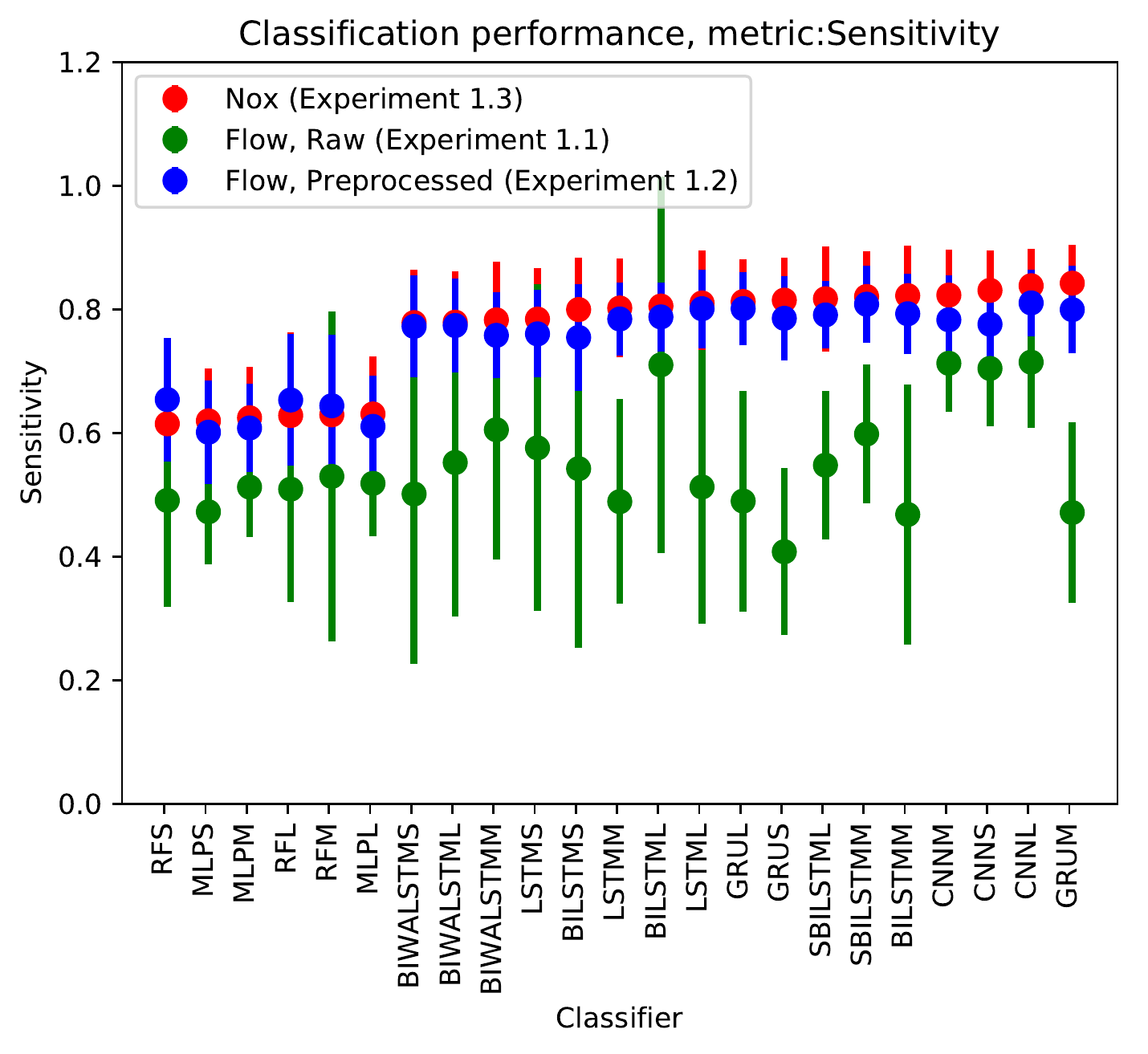}\\(b)}&
     \makecell{
       \includegraphics[width=0.30\textwidth]{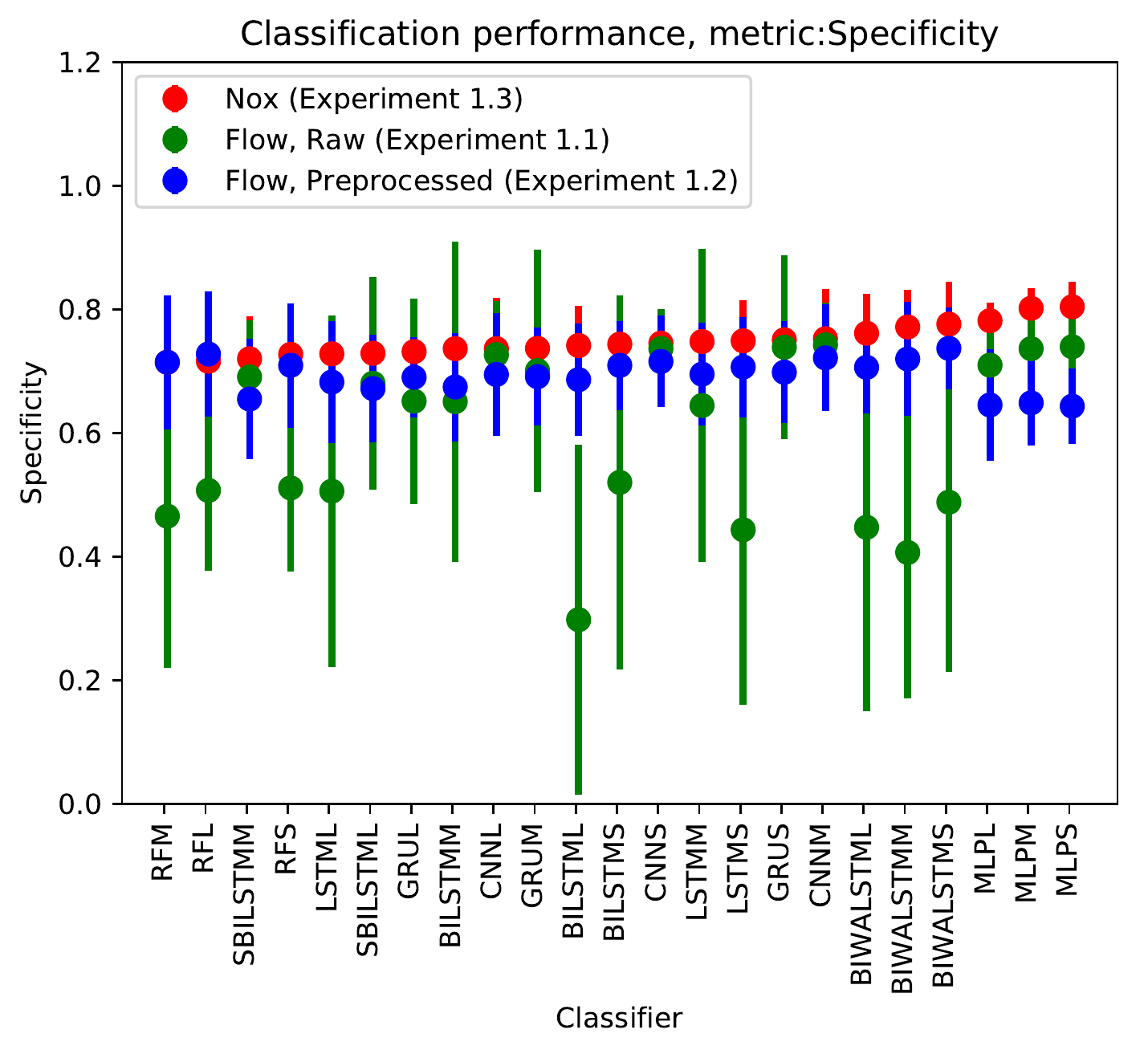}\\(c)}
     \end{tabular}
     \caption{Accuracy, sensitivity, and specificity for Experiment 1.}
    \label{figure:exp1accsensspec}
\end{figure*}

The $\kappa$ results for Experiment 1 are presented in Figure \ref{figure:exp1}, and the accuracy, sensitivity, and specificity in Figure \ref{figure:exp1accsensspec}.

On average, BLA leads to an increase in $\kappa$ of 0.2889$\pm$0.1091 points for any given classifier. CNN classifiers benefit far less from BLA than the others, with a mean increase in $\kappa$ of only 0.1263$\pm$0.1054. This indicates that CNN is resilient to baseline shifts and wandering. Overall, sensitivity is slightly higher than specificity, indicating that the classifiers are better at correctly classifying apneic than non-apneic periods. Deep NN-based classifiers substantially outperform the older RF and MLP classifiers. The highest $\kappa$ of 0.5062$\pm$0.0575 is obtained with BIWALSTMS and BLA, closely followed by CNNM with a $\kappa$ of 0.5052$\pm$0.0485. Surprisingly, RF yields better results with Flow data than with Nox data. Since RF nevertheless yields the overall lowest performance for all metrics and data sets, this result is likely a result of RF failing to properly learn the important features of the Nox data. BLA clearly increases accuracy substantially, and has an equalizing effect on sensitivity and specificity. The latter is particularly prominent for MLP and GRU classifiers. The highest accuracy is obtained with CNNM and BLA, i.e., 0.7609$\pm$0.0284, with a sensitivity of 0.7833 and a specificity of 0.7217. 

We see that the SD are much larger than the differences in mean values. We can therefore not conclude from the mean $\kappa$ alone that BIWALSTM is the Superior classifier. We must also consider the amount of resources required for training and inference. We measure the average time spent training and testing per epoch. Training involves inference and back-propagation of four batches, and testing the inference of one batch\footnote{This batch is smaller than 1000 periods, since there are only 598 periods in the test set. Similarly, the final batch in the training set consists of only 766 periods.}. The measurement from the first epoch is excluded since it includes time spent on initialization which is not relevant in this investigation. We find a substantial difference among the classifiers. Recurrent networks spend a factor of 2.40 to 77.46 more time per epoch for training than feed-forward networks. The corresponding numbers for testing time is 3.57 to 41.59, which involves the inference of one batch. BIWALSTM is by far the most resource intensive, requiring just above 224 milliseconds per epoch for training and about 18 milliseconds per epoch for testing. CNNM, which obtain the same classification performance, only requires an average of 33 milliseconds and 2 milliseconds per epoch for training and testing, respectively. These differences can translate into differences in waiting times of hours vs. days with large data sets or models.

\section{Mixing High- and Low-Quality Data}
The data from the Nox-T3 sensor is of much higher quality than that from the Flow sensor. It is well-known that both data quality and quantity affects classification performance. We hypothesize that mixed training with Flow and Nox data can improve the accuracy of classifying Flow data. Similar approaches have previously shown to yield promising results in other domains, e.g., in the field of transfer learning \cite{pan2009survey}. In Experiment 2, we investigate this hypothesis. Interestingly, the results from Experiment 2 suggest that classifiers that are pre-trained on high-quality data can be used \emph{as-is} to classify data data from different sensors of lower quality, and still obtain a performance comparable to that of training on data from the target, low quality sensor. This has significant implications, since the collection of large amounts of data is expensive and may not be possible for many producers of inexpensive home electronics due to privacy and resource restrictions. The results furthermore suggest that the lack of large-scale, high-quality data sets may in some cases be compensated for by collecting a much smaller amount of data from high- and low-quality sensors at the same time. We show that this results in substantial savings of training time.

We perform four sub-experiments 2.1-2.4. Each sub-experiment is repeated 10 times. Due to the large number of repetitions, the total number of experiments is very large. For this reason, and because we focus on the potential improvements beyond what was achieved in Experiment 1.2, we use the CNNM classifier that achieves the overall best performance in Experiment 1.2. In Experiment 2.1, we train the classifier on $N$ and test it on $FB$. This reflects a scenario where a classifier trained on high-quality data is made available and used \emph{as-is} to classify low-quality data. Experiments 2.2 to 2.4 are trained with a mix of Flow and Nox data. In Experiment 2.2, the training is performed \emph{sequentially}, i.e., first with data from the Nox-T3 then with data from the Flow sensor. A complete 10-fold CV training procedure is first completed with $N$. The resulting pre-trained models (one per fold) are subsequently used as the initial models for a complete 10-fold CV run using $FB$ as both training and test data. The results are relevant for scenarios where a classifier trained on high-quality, closed data from, e.g., clinical studies is made available for further training on readily available low-quality data. Due to catastrophic forgetting \cite{french1999catastrophic}, the final classifier in Experiment 2.1 may no longer be able to accurately classify the high-quality Nox-data, only the low-quality Flow data it is intended for. In Experiments 2.3 and 2.4, training is performed \emph{simultaneously} with Nox-T3 and Flow data. In Experiment 2.3, each of the training sets produced during 10-fold CV is concatenated with $NF$, and only $FB$ is used for testing. This configuration is found in scenarios where both high- and low-quality training data is available from the same set of individuals. Experiment 2.4 is identical to 2.3, except $NF$ is replaced with $N$. This gives an estimate of performance obtainable in scenarios where high-quality data is available from many more individuals than for low-quality data. We compare with results from Experiments 1.2 and 1.3 repeated 10 times where we use only high- and low-quality data from Experiment 1.2 and 1.3 repeated 10 times.

In Experiments 2.2, 2.3., and 2.4, Flow and Nox data can be weighted differently during training. This is achieved by multiplying the loss of samples in the Nox-data by $w \in [0,1.0]$ before they are used for back-propagation. We execute these sub-experiments multiple times with varying $w$ to estimate the optimal weight and to study the relative impact of low- and high-quality data during training.

\begin{figure}
  \centering
  \includegraphics[width=0.75\textwidth]{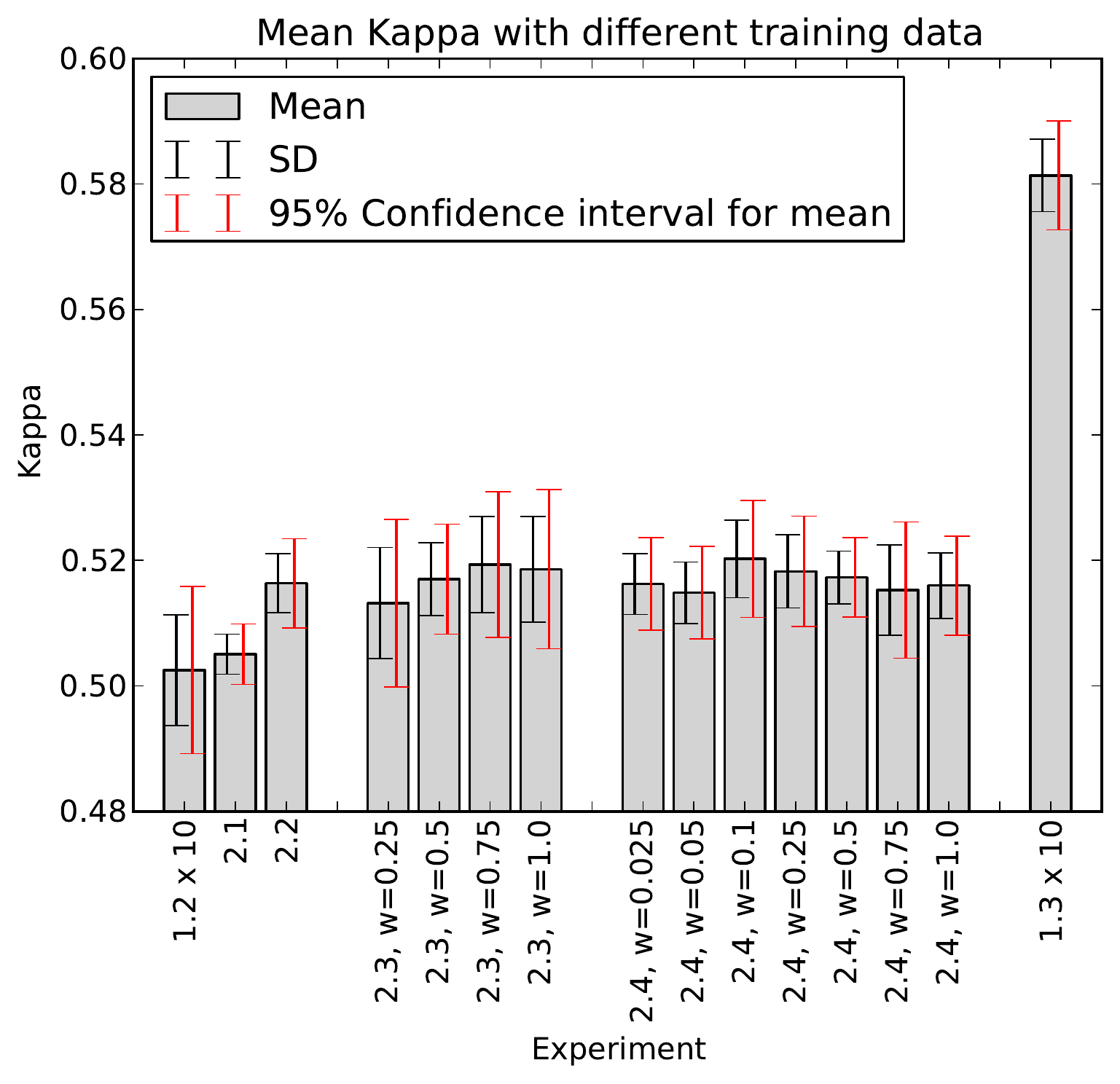}
  \caption{Results from Experiment 2. $w$ is the weight by which the loss from samples in Nox data is multiplied before back-propagation.}
  \label{fig:mixed}
\end{figure}

The results from Experiment 2 are presented in Figure \ref{fig:mixed}. The x-axis shows the different sub-experiments, and the y-axis the mean $\kappa$ for CNNM. This is not the mean across folds, as in Experiment 1. It is the mean across the results from 10 different repetitions of the complete 10-fold CV procedure with different random seeds. The result from each such repetition is the mean across the 10 folds. This makes visible the variance resulting from different initializations of the weights in the NN. When the training set consists of both Nox and Flow data, the weight on the Nox data is denoted as $w$ in the labels on the x-axis.

The results show an overall improvement in classification performance when using a mix of Nox and Flow data for training. The $\kappa$ from Experiments 2.2-2.4 are between 0.0107 and 0.0177 higher than in Experiment 1.2. While the mean accuracy in Experiment 1.2 is 0.7580, the mean accuracies in Experiments 2.2-2.4 are 0.7650, 0.7647, and 0.7647, respectively. There is no clear difference in $\kappa$ among Experiments 2.2-2.4, since differences in the mean values are much smaller than the corresponding SD. This suggests that it does not matter much for the performance: (1) If we first train on Nox data then on Flow data (Experiment 2.1), or train on both in parallel (Experiments 2.2-2.4), (2) If we use the complete A3 data set (Experiment 2.4) or only the sub-set obtained simultaneously with the Flow data (Experiment 2.3), (3) How the Nox and Flow data is weighted during training. In Experiment 2.3, the training benefits from the pre-liminary synchronization of Nox- and Flow-data. In Experiment 2.4, training benefits from the large amount of training data available. The similarity of the results from these two experiments suggest that these two factors have a similar positive impact. This means that using the complete, large-scale Nox-T3 data from the A3 study yields a similar performance as using the much smaller sub-set that was collected at the same time as the Flow data. Due to the high SD and 95\% CI we cannot rule out the existence of small differences between these two approaches that would be uncovered with a larger data set.

We obtain similar $\kappa$ in Experiment 2.1 as in 1.2, i.e., in the absence of Flow data in the training set. This result suggests that classifiers that are pre-trained on high-quality data can be used \emph{as-is} to classify data from different sensors of lower quality, and still obtain a performance comparable to that of training on data from the target, low quality sensor.

A proper comparison among the approaches requires to consider the training times. The training data used in Experiments 2.3 is only twice as big as that used in Experiment 1.2. The mean training time per epoch is only about twice as big, i.e., 71 milliseconds. The data set used in Experiments 2.1, 2.2, and 2.4 use the complete A3 data set for training, which comprises 232 batches compared to only four in Experiment 1.2. The mean training times are significantly larger, i.e., 1708 milliseconds per epoch. While Experiments 1.2 and 2.3 complete in minutes, 2.1, 2.2, and 2.3 require several hours to complete. This large difference in training time implies that it is preferable to use $NF$ instead of $N$ to augment the training data from $F$.

\section{Per-Subject SA Severity Estimation}
\label{section:exp3}
Experiment 1 and 2 are designed with comparability with related works in mind. By balancing the data set, these experiments can provide a good indication of the performance of the classifiers in the general case, i.e., without any assumption of SA prevalence. The goal of Experiment 3 is to study how accurately we can estimate the SA severity of a subject. Our results show a clear correlation between actual (from manual annotation of Nox-T3 data) and predicted (from ML on Flow data) AHI, albeit naturally not as strong as between, e.g., AHI from PG and PSG. Such AHI estimation precludes balancing the test set, causing the class distribution of the test set to be determined by the prevalence of apneic periods. A key insight from Experiment 3 is that, as a result, the decision of whether or not to balance the \emph{training set} has a significant impact on results, the degree of which depends on the prevalence of SA. Careful attention should therefore be given to properly align this decision with the requirements of the particular screening context. We find that balanced training data leads to fewer false negatives which is generally preferable from a purely medical perspective. But it leads to a consistent over-estimation of SA severity, and may not be preferable from a large-scale, societal perspective, due to the potential of added cost of incorrectly prescribed expensive medical examinations. Conversely, imbalanced training data results in an overall higher prediction accuracy at the cost of a higher number of false negatives. We show that a compromise can be made by averaging the predictions made with balanced and imbalanced data. Averaged predictions, and those from imbalanced data, are better than those from balanced data for discriminating between subjects with AHI above or below 15. This can have significant implications in practice, as an AHI-threshold of 15 is commonly used as a guideline for prescribing a PSG examination.

We obtain these findings by changing the CV procedure compared to Experiments 1 and 2. It is only possible to estimate the SA severity of an individual by analyzing the entire data set from this individual. For each iteration in the CV procedure, the test set consists of all the recordings of a given subject, while the training set consists of the remaining recordings. To study the effect of balancing the training set, we compare results with classifiers trained on balanced ($C_b$) and imbalanced data ($C_i$). In addition, we create the combined classifier $C_c$ by averaging the predictions from $C_b$ and $C_i$. These classifiers use the best-performing CNNM architecture in Experiment 1. 

\begin{figure}
  \centering
  \includegraphics[width=0.75\textwidth]{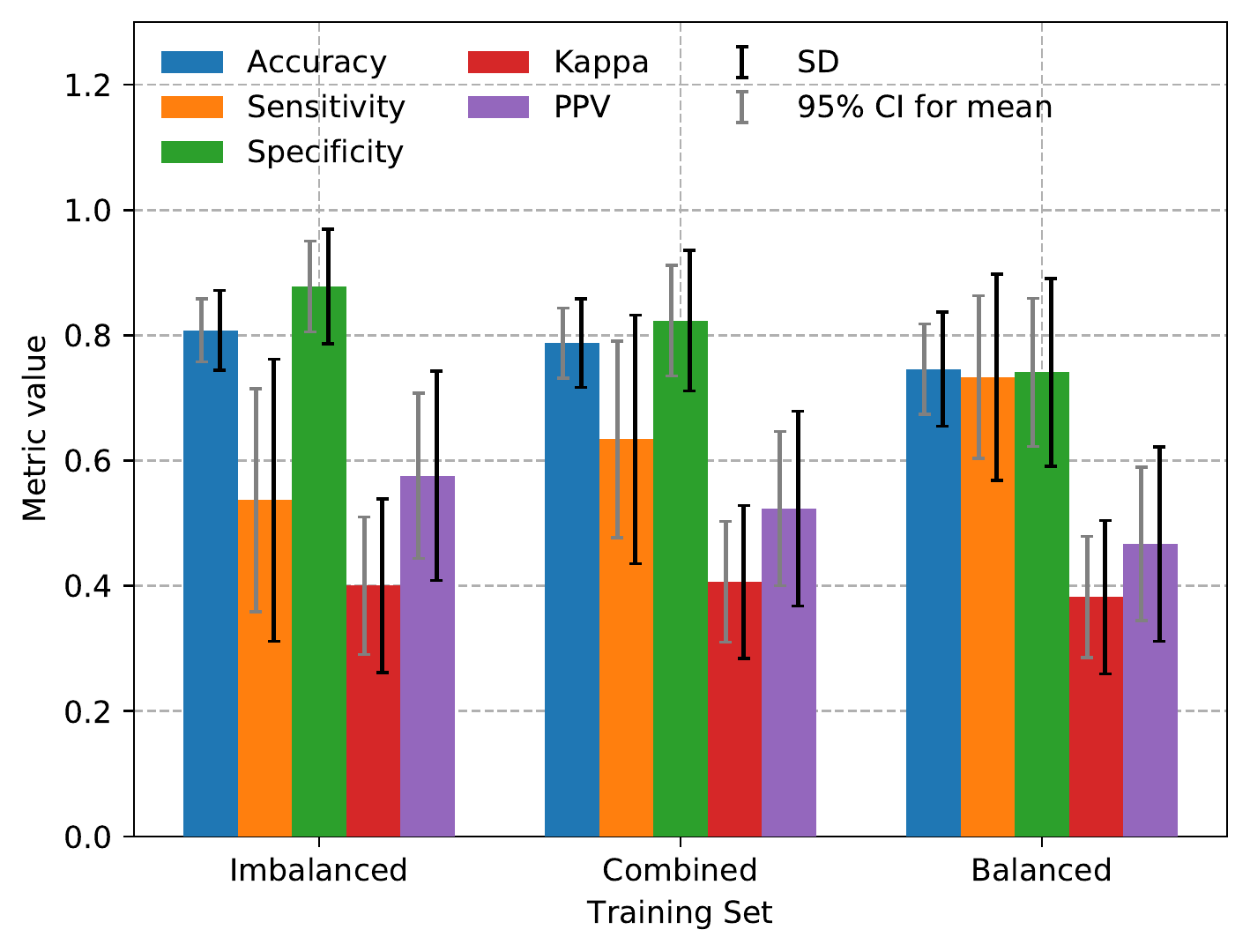}
  \caption{Results from Experiment 3.}
  \label{fig:combined}
\end{figure}

A proper understanding of classification performance with imbalanced data requires a joint assessment of several metrics \cite{he2009learning}. Since Experiment 3 focuses on SA-severity estimation, most of the results are based on comparisons between \emph{predicted} and \emph{actual AHI}. We compute actual AHI by counting the actual number of disrupted breathing events in the Nox-data\footnote{Note that this AHI differs from the actual AHI of the subjects presented in Section \ref{section:dataaquisition}, since a significant portion of the Nox-data is removed during pre-processing (see Section \ref{section:preprocessing}).}. Predicted AHI is computed based on the total number of predicted apneic periods. Since the number of apneas/hypopneas may differ from the number of apneic periods, the results in Experiment 3 reflect the negative performance impact of dividing the data into periods. This is not the case in most related works since online data sets often only include per-period labels \cite{mostafa2019systematic}. As a result, Experiment 3 yields a lower, but more realistic performance.

\begin{figure*}
  \centering
  \begin{tabular}{ccc}
     \makecell{
       \includegraphics[width=0.30\textwidth]{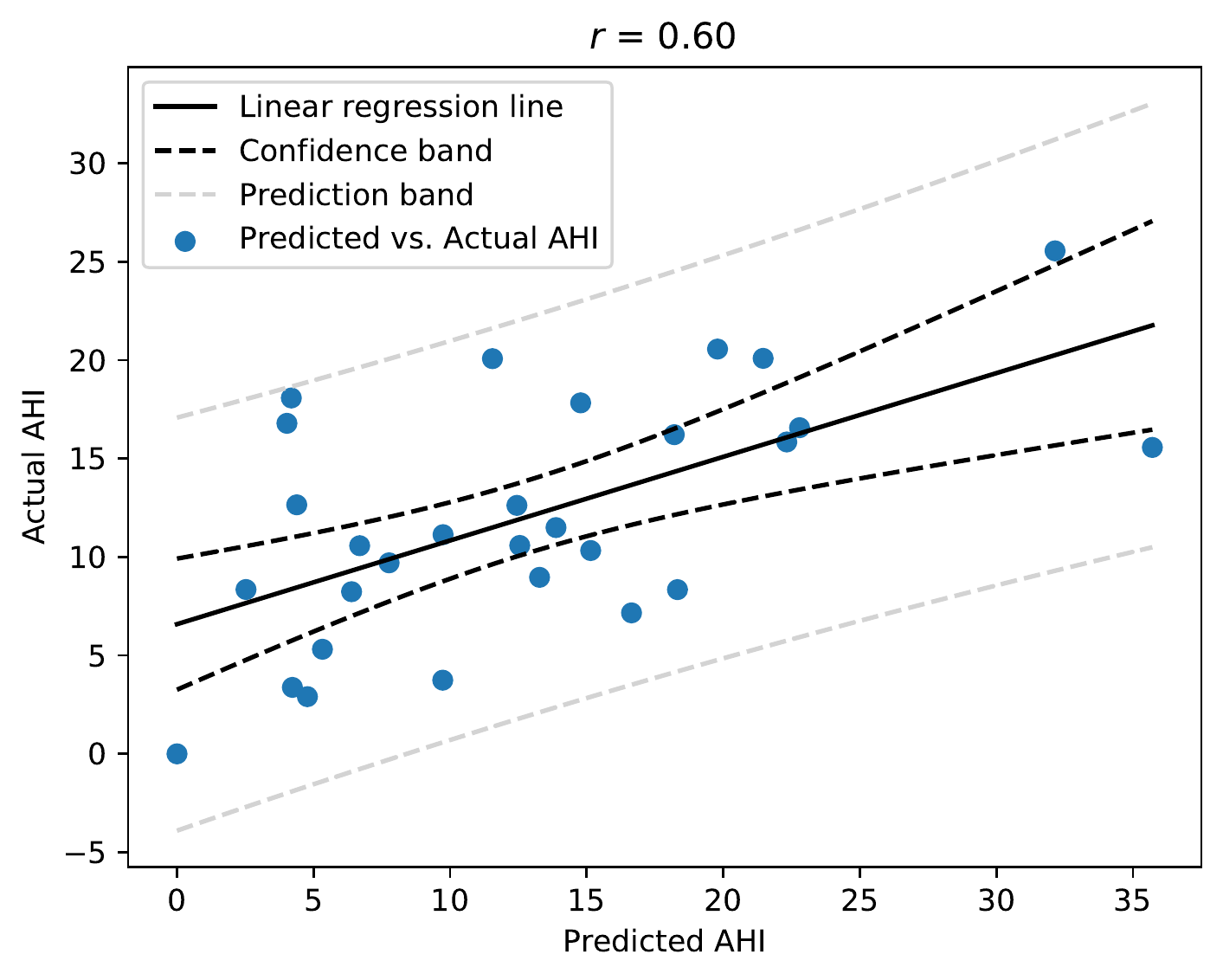}\\(a) $C_i$.}&
     \makecell{
       \includegraphics[width=0.30\textwidth]{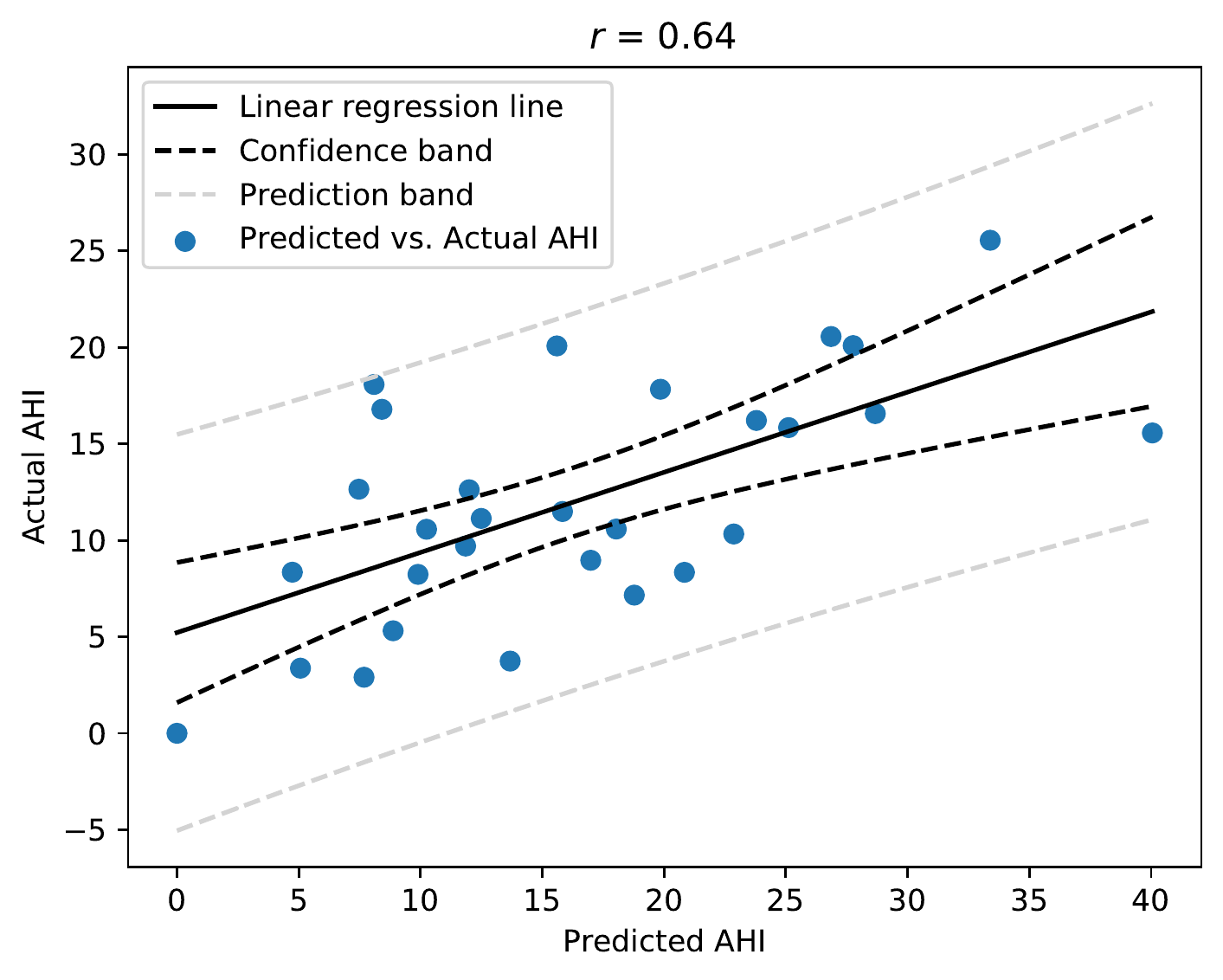}\\(b) $C_c$.}&
     \makecell{
       \includegraphics[width=0.30\textwidth]{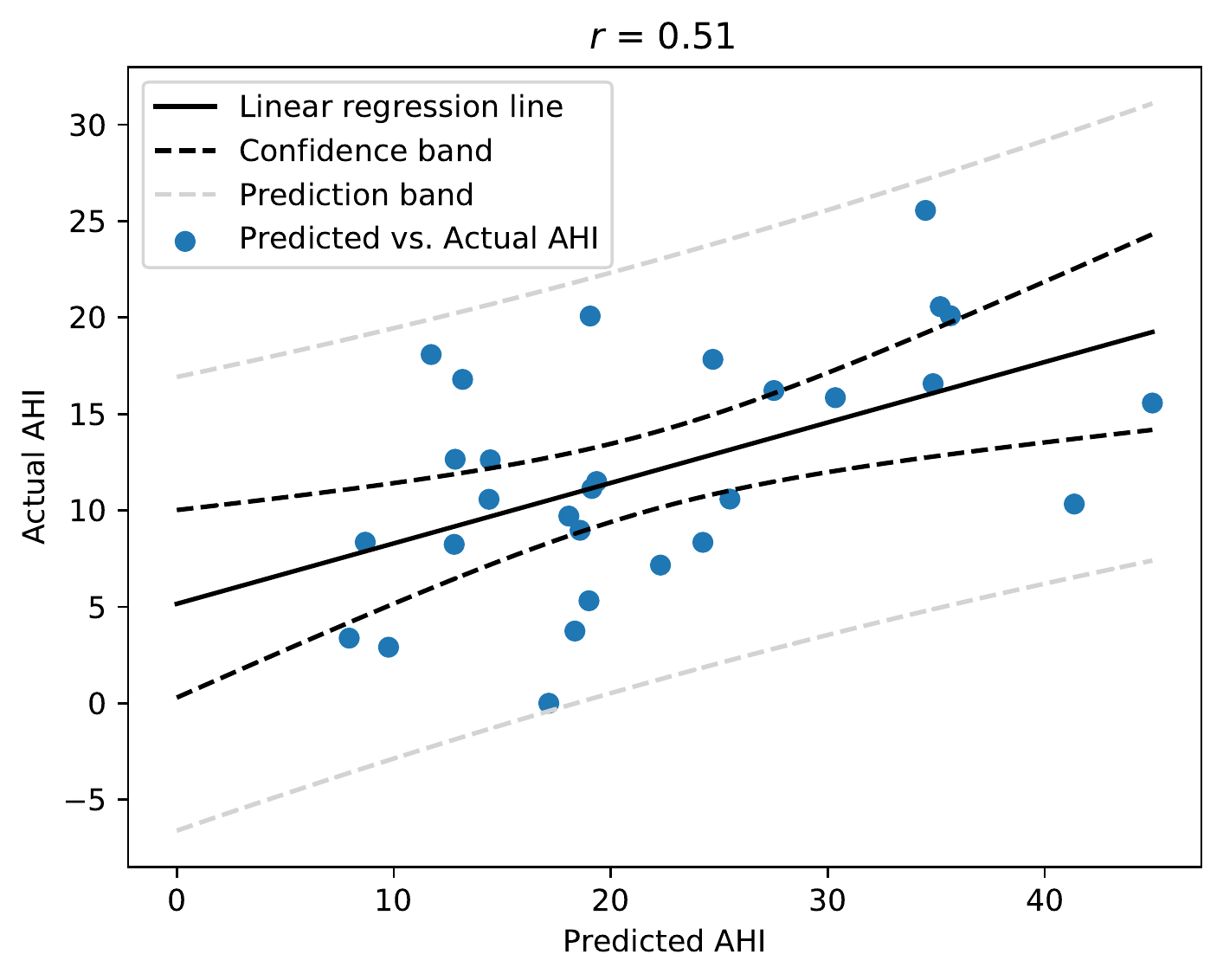}\\(c) $C_b$.}\\
     \makecell{
       \includegraphics[width=0.30\textwidth]{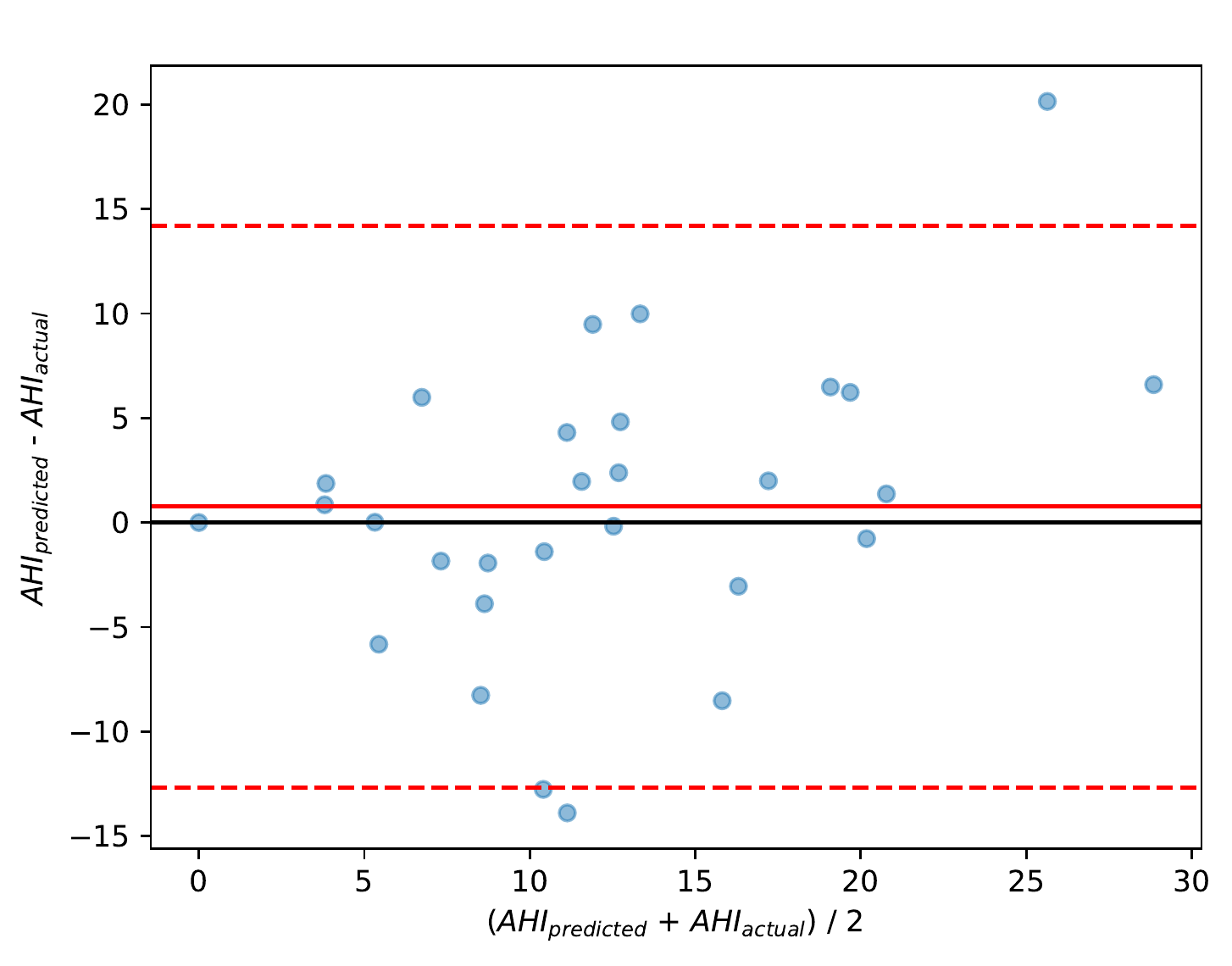}\\(d) $C_i$.}&
     \makecell{
       \includegraphics[width=0.30\textwidth]{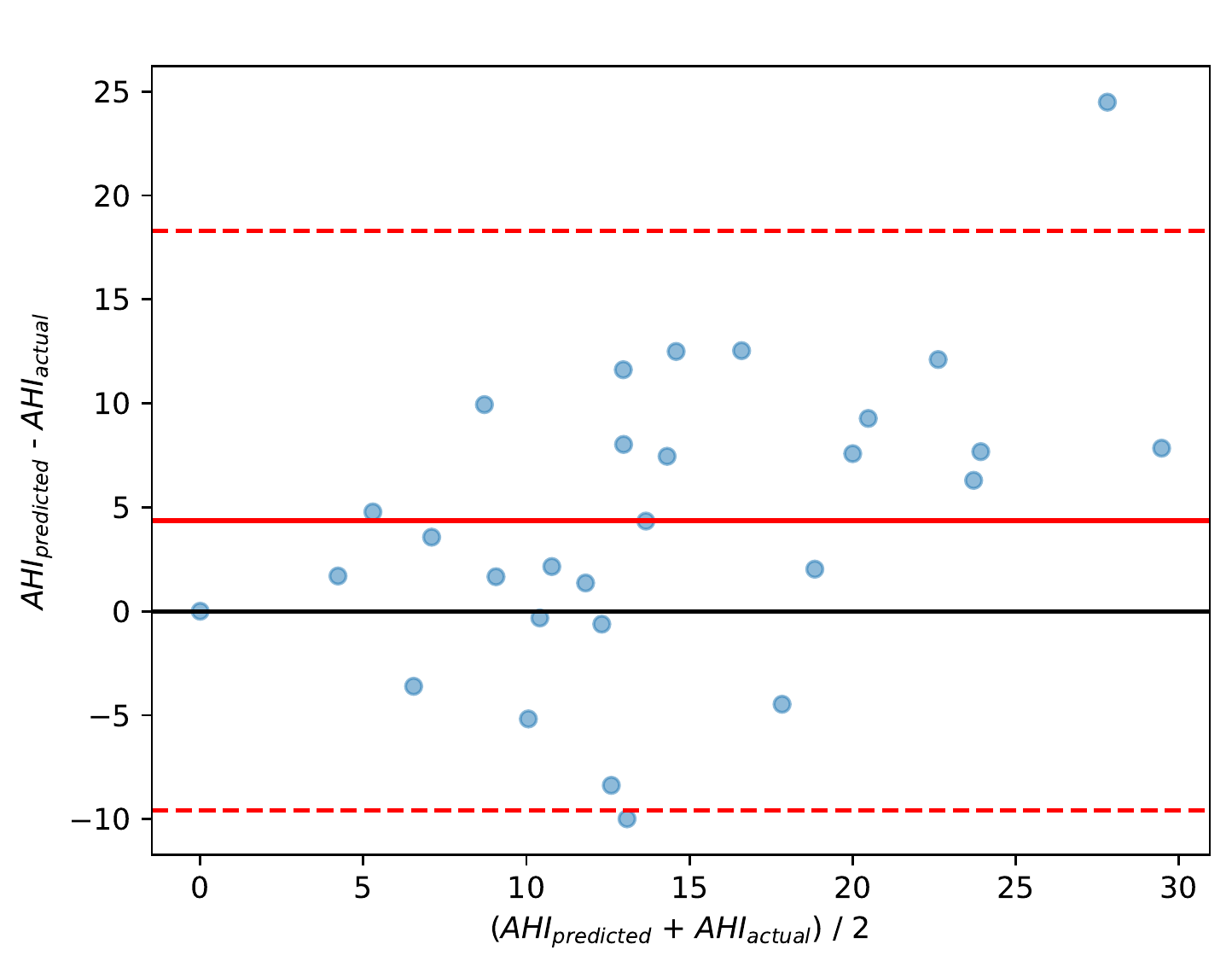}\\(e) $C_c$.}&
     \makecell{
       \includegraphics[width=0.30\textwidth]{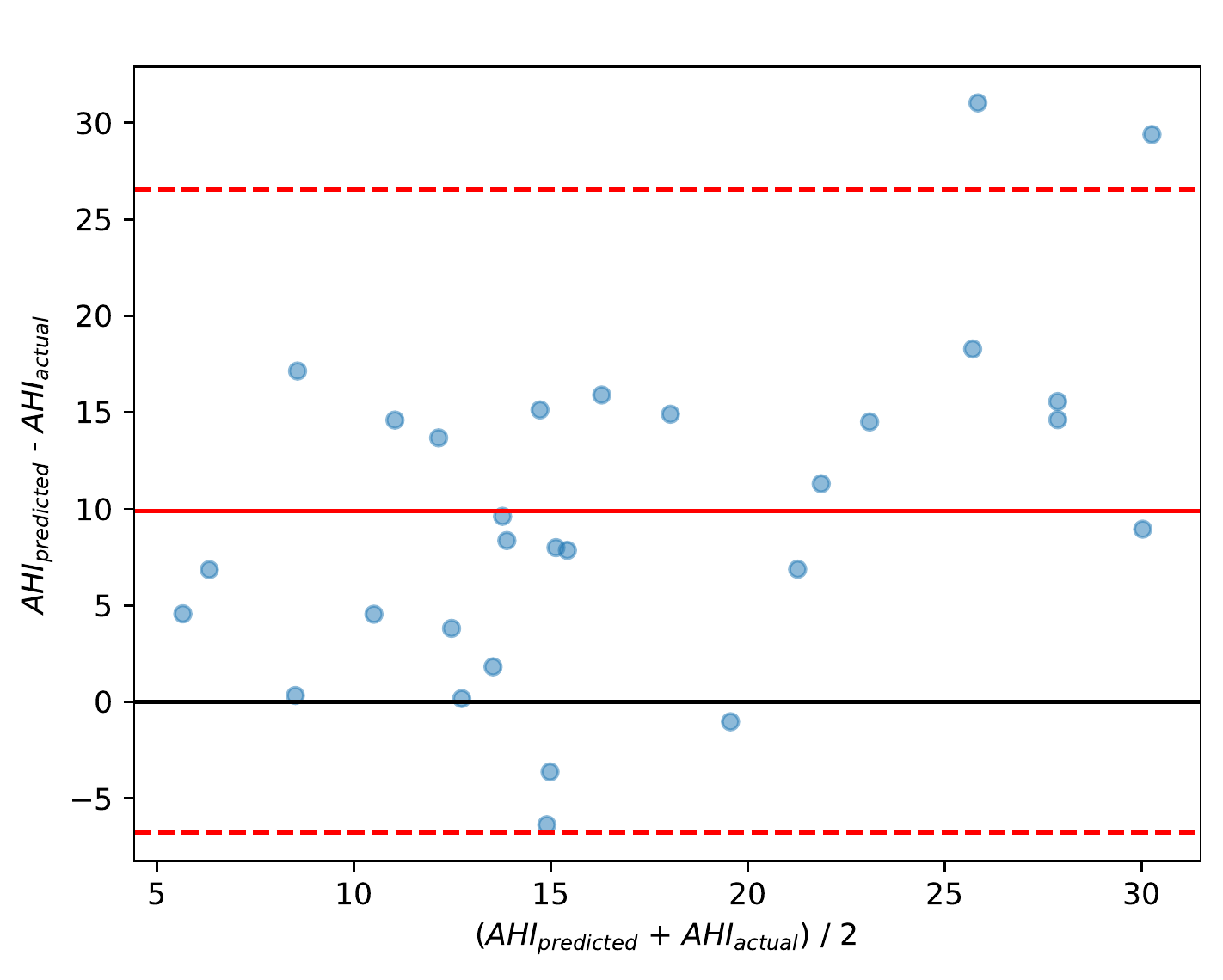}\\(f) $C_b$.}\\
  \end{tabular}
  \caption{Scatter (a-c) and Bland-Altman plots (d-f) for Experiment 3. The results are with $C_i$, $C_c$, and $C_b$ from left to right, respectively.}
  \label{figure:exp3scatter}
\end{figure*}

\begin{figure*}
  \centering
  \begin{tabular}{ccc}
     \makecell{
       \includegraphics[width=0.30\textwidth]{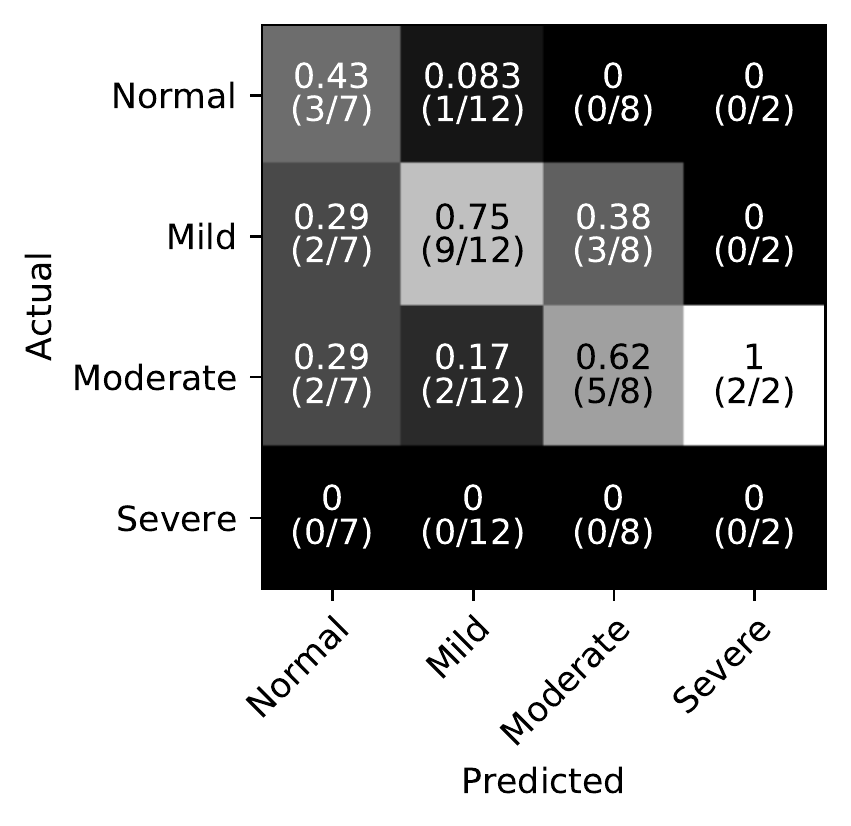}\\(a) $C_i$.}&
     \makecell{
       \includegraphics[width=0.30\textwidth]{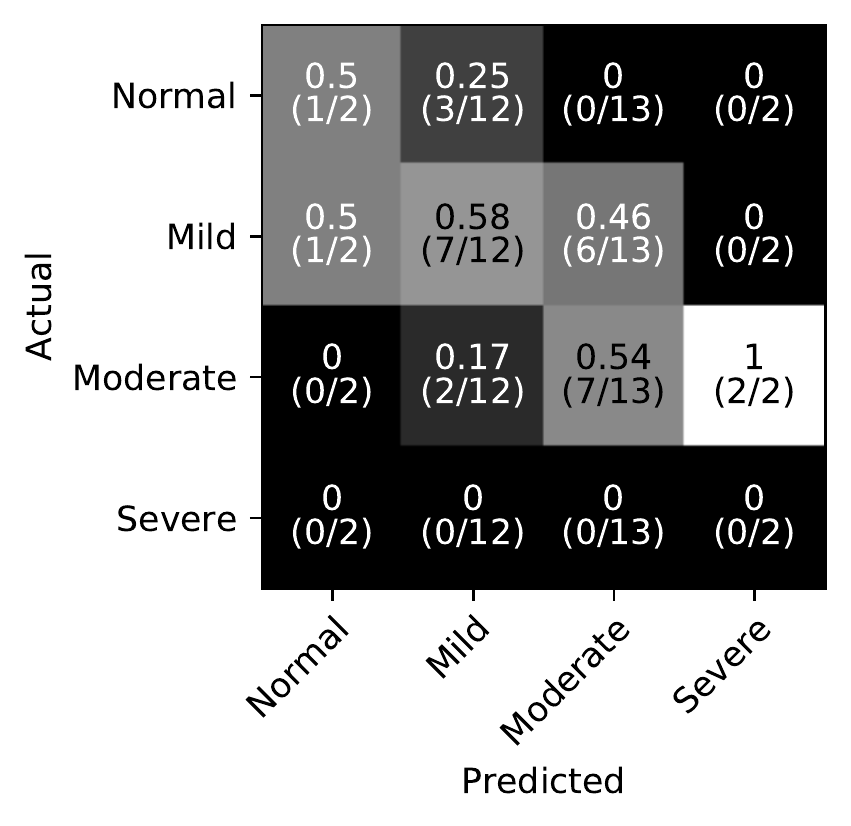}\\(b) $C_c$. }&
     \makecell{
       \includegraphics[width=0.30\textwidth]{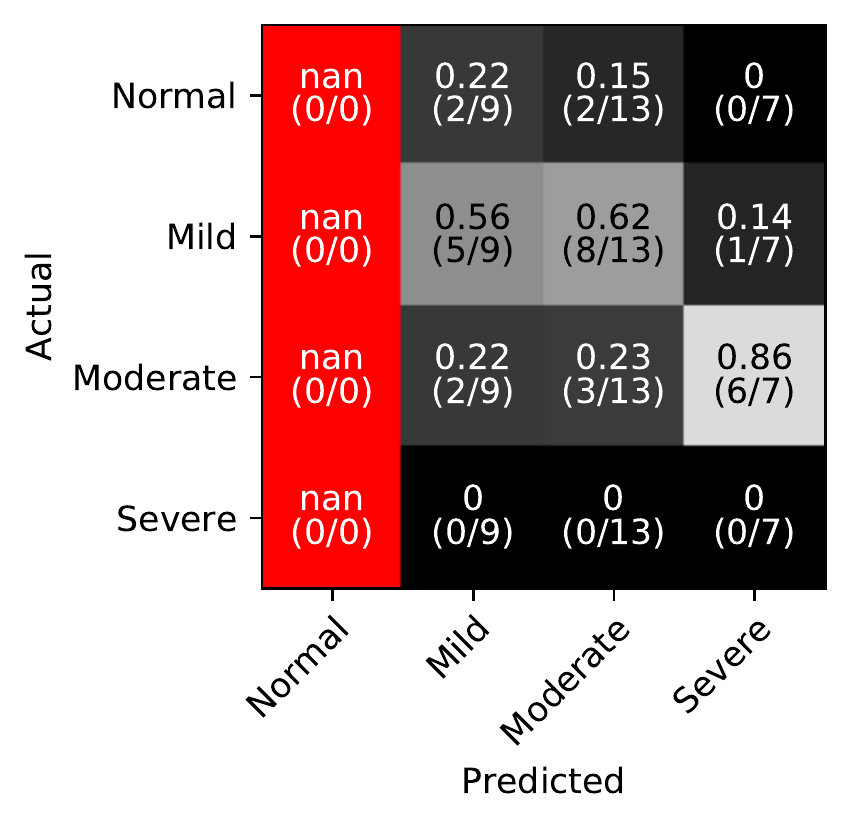}\\(c) $C_b$.}
  \end{tabular}
  \caption{Confusion matrices for Experiment 3 with the fraction of individuals in predicted SA severity groups that were actually in that severity group.}
    \label{figure:exp3confusion}
\end{figure*}

\begin{table*}
  \resizebox{\textwidth}{!}{
    \centering
    \begin{tabular}{ l l | l | l | l | l || l | l | l | l | l || l | l}
      &&\multicolumn{2}{c|}{\textbf{AHI $<$ TH}}&\multicolumn{2}{c||}{\textbf{AHI $\ge$ TH}}&\multicolumn{5}{c}{}\\
      &\textbf{Threshold}&\textbf{Act.}&\textbf{Pred.}&\textbf{Act.}&\textbf{Pred.}&\textbf{Acc.}&\textbf{Sens.}&\textbf{Spec.}&\textbf{$\kappa$}&\textbf{PPV}&\textbf{P(A)}&\textbf{P(A$\vert$P)}\\
      \hline
      &5&4&7&25&22&0.8276&0.8400&0.7500&0.4487&0.9545&0.8621&0.9545\\
      \textbf{$C_i$:}&15&18&19&11&10&0.7586&0.6364&0.8334&0.4781&0.7000&0.3793&0.7000\\
      &30&29&27&0&2&0.9310&NaN&0.9310&0.0000&0.0000&0.0000&NaN\\
      &&&&&&&&&&&\\
      &5&4&0&25&29&0.8621&1.0&0.0&0.0&0.8621&0.8621&0.9259\\
      \textbf{$C_b$:}&15&18&9&11&20&0.5517&0.8182&0.3889&0.1786&0.4500&0.3793&0.6000\\
      &30&29&22&0&7&0.7586&NaN&0.7586&0.0000&0.0000&0.0000&NaN\\
      &&&&&&&&&&&\\
      &5&4&2&25&27&0.8621&0.9600&0.2500&0.2658&0.8889&0.8621&0.8276\\
      \textbf{$C_c$:}&15&18&14&11&15&0.7241&0.8182&0.6667&0.4528&0.6000&0.3793&0.4500\\
      &30&29&27&0&2&0.9310&NaN&0.9310&0.0000&0.0000&0.0000&NaN\\
  \end{tabular}}
  \caption{Results for Experiment 3. Performance of binary classification with thresholds corresponding to the different SA severity classes. P: Predicted positive. A: Actual positive. TH: AHI Threshold. NaN: Not a Number (incomputable).}
  \label{table:binarySC}
\end{table*}

As a basis to understand the accuracy of estimating the AHI of individuals, we first investigate how accurately the 60-second periods are classified for each individual. The mean, SD, and CI across all individuals are shown in Figure \ref{fig:combined}. All remaining results are based on comparisons between the actual AHI of an individual and the AHI estimated using the classified periods for that individual. Figure \ref{figure:exp3scatter} shows scatter plots (a-c) and Bland-Altman plots (d-f) that compare actual and predicted AHI per individual. The scatter plots include the 95\% confidence and prediction bands. These outline the 95\% confidence intervals for the mean and individual AHI for the ground truth given predicted AHI, respectively. Figure \ref{figure:exp3confusion} shows the confusion matrices for SA severity groups. They denote the probability of a subject belonging to a certain actual severity group given a predicted group. According to the balanced predictions, no subjects belong to the severity group \emph{mild}, i.e., these values cannot be computed. The actual number of subjects in severity groups normal, mild, moderate, and severe is four, 14, 11, and zero, respectively. Table \ref{table:binarySC} shows the classification performance where subjects are divided into only two groups, i.e., with AHI above or below the thresholds that separate severity groups. Since no subjects belong to the actual severity group \emph{severe}, the corresponding sensitivity, $\kappa$, and PPV values are either not computable or not meaningful. The $P(A)$ column presents the prior probability of a positive based on the SA prevalence among the 29 subjects, and the $P(A|P)$ column presents the posterior probability of an actual positive given a positive test as given by Bayes' formula $P(A \vert P)=\frac{P(P \vert A)P(A)}{P(P)}$. The difference between the two right-most provides an estimate of how much a positive test increases the probability of an actual positive that accounts for SA prevalence (as opposed to PPV).

Figure \ref{fig:combined} shows that $C_i$ outperforms $C_b$ for all metrics except sensitivity. $C_c$ performs somewhere in between $C_i$ and $C_b$. These findings are a result of the imbalanced test set, which has an over-representation of normal periods. This problem of \emph{intrinsic imbalance} is common in the biomedical domain \cite{he2009learning}, and the smaller the data set is, the larger its impact \cite{weiss2001effect}. This configuration leads to different biases in the training set which is learned by the classifier: a balanced training set has a very different class distribution than the test set, while an imbalanced training set contains many more normal than apneic periods. $C_i$ benefits in terms of accuracy and specificity from the much higher number of non-apneic (10117) than apneic (2988) periods. $C_b$ has seen an equal number of apneic and non-apneic periods during training, while $C_i$ has seen many more non-apneic ones. Therefore, $C_i$ more accurately classifies non-apneic periods, which are much more numerous in the test set than apneic ones. This explanation is confirmed by the fact that $C_b$ outperforms $C_i$ on the balanced validation set in terms of all metrics except accuracy and specificity. These results should be seen in light of the relatively large CI, indicating that a larger data set is desirable to increase the confidence in the mean values.

Figures \ref{figure:exp3scatter} (a-c) show that we obtain r-values between 0.51 and 0.64 with different prediction approaches. Interestingly, $C_c$ outperforms $C_b$ and $C_i$, in terms of the r-value and the magnitude of the prediction and confidence intervals. There is a clear correlation between predicted and actual AHI. As expected, the r-values are lower than what can be obtained with manual scoring of data from clinically certified PG equipment. For instance, in \cite{chang2019validation} an r-value of 0.791 is obtained when comparing manual scoring of Nox-T3 and PSG data. This is explained by the results in Figure \ref{fig:combined}. Values for sensitivity show that only 53.3\% to 73.3\% of the apneic periods for each subject are correctly identified as such. We see that the slope of the regression line is lower than 1, which means that predicted AHI are more dispersed than the actual AHI. The fact that the performance in our results is not as good as previous works with clinically certified equipment is not surprising. A major reason is the much lower quality of data from consumer electronics like the Flow sensor. While recent deep NN are known to excel in classification even of noisy data, the presence of irreducible errors in both data and labels imposes an upper limit on the achievable classification performance. This limit is studied in literature both theoretically and empirically \cite{tumer1996estimating, hestness2017deep}. Furthermore, a large amount of data is required to achieve this upper performance limit \cite{kristiansen2020machine}. Intuitively, the amount of data required increases with decreased data quality. The Flow data is both of lower quality than the Nox data and much smaller, i.e., it consist of only 49 recordings from 29 subjects with an average length of 4.46 hours of data. With this few subjects, personal differences have a non-neglectable impact on the ability of the classifiers to generalize.

The BA plots show that $C_b$ is biased towards higher SA severity groups, which is not the case for $C_i$. The effect is reflected in the confusion matrices in Figure \ref{figure:exp3confusion}. While $C_i$ most often correctly predicts the SA severity group, $C_b$ tends to over-estimate SA severity. $C_b$ has a higher sensitivity and a lower PPV than $C_i$ (see Figure \ref{fig:combined}), causing $C_b$ to classify more periods as apneic on average.

Table \ref{table:binarySC} reflects the results above. Since $C_b$ over-estimates SA severity, only half of the subjects are correctly classified at $TH=15$, and only 38.89\% of negative subjects (AHI $<$ 15) are classified as such. $C_c$ yields good results for $TH=15$, i.e., (1) a relatively high accuracy for $TH=15$, (2) the same sensitivity as with $C_b$, and (3) a relatively modest decrease in specificity compared to $C_i$. We see that $C_i$ outperforms the others for all thresholds (TH) in terms of accuracy, specificity, $\kappa$, and PPV, with the exception of $TH=5$ for which $C_c$ achieves a slightly higher accuracy. $C_b$ and $C_c$ outperform $C_i$ in terms of sensitivity. Notice that $TH=15$ yields the lowest accuracy but the highest $\kappa$ among all TH. This is explained by the relatively balanced class distribution at this threshold, i.e., 62.07\% of the subjects have an actual AHI above this threshold. With $TH=5$ and $TH=30$, 86.2\% and 0\% are above the threshold, respectively. Such imbalances result in artificially high accuracies. The usefulness of a test depends on how much a positive test result increases the probability an actual positive (relative to the prior probability). The columns $P(A)$ and $P(A|P)$ show that $C_i$ leads to the largest increase for all TH. Importantly, with $TH=15$, a positive test result almost doubles the probability of SA from 37.93\% to 70.00\%.

\section{Inference on a Smartphone}
\label{section:exp4}
In Experiment 4, we investigate whether the classifiers can in practice be used in our envisioned target scenario, i.e., on a patient's smartphone. We assume that a classifier is trained on a powerful computer and thereafter made available for use at home. Smartphones might be ideal for this purpose because nearly everyone owns one, but are severely resource limited compared to desktop or cloud computers. We use Tensorflow Lite to convert the classifiers into \emph{float models} that are suitable for mobile devices. Our results show that it can analyze the data from a complete night mostly within one second on a smartphone from 2012, i.e., the Samsung Galaxy Tab S3. Surprisingly, this is achievable with almost no reduction in classification performance. The classifier is well-suited for use on a smartphone.

The float models are constructed to be smaller than the original models without a significant loss in prediction performance, and can be executed using the low-overhead \emph{tflite} operations in Tensorflow Lite. The model can be \emph{quantized} to reduce inference times further, but at the cost of numerical precision and classification performance. We measure the inference times of both Float and quantized versions of the CNN models from Experiment 1. We measure the time spent classifying data from one complete night, as this reflects the waiting time required for analysis the morning after a night of data collection. To study the impact of model size, we compare results from CNNS, CNNM, and CNNL. After converting the trained models, they are used for inference on a Samsung Galaxy Tab S3 with a 1.6GHz quad-core Qualcomm Snapdragon 820 processor and 4GB of RAM, running the Android 7.0 operating system and using Tensorflow version 2.2.0. Since the main focus of Experiment 4 is inference efficiency and the impact on model reduction on performance, 10 training epochs was sufficient to obtain representative results.
\begin{table}
  \centering
  \begin{tabular}{ l | c | c | c | c | c }
    \emph{Classifier}&\emph{Parameters}&\emph{Size}&\emph{Inf. time}&\emph{$\kappa$}&\emph{Acc.}\\
    \hline
    \emph{CNNS:}&&&&&\\
    Original&166.658&2-2.2MB&&0.4903&0.7512\\
    Float&166.658&654KB&0.55ms&0.4989&0.7525\\
    Quantized&166.658&168KB&0.34ms&0.4928&0.7492\\
    &&&&&\\
    \emph{CNNM:}&&&&&\\
    Original&413.058&5-5.2MB&&0.5052&0.7609\\
    Float&413.058&1617KB&1.09ms&0.4820&0.7458\\
    Quantized&413.058&411KB&0.55ms&0.4882&0.7492\\
    &&&&&\\
    \emph{CNNL:}&&&&&\\
    Original&1.069.186&12.8-13.1MB&&0.5020&0.7554\\
    Float&1.069.186&4181KB&1.93ms&0.4674&0.7375\\
    Quantized&1.069.186&1055KB&0.93ms&0.4531&0.7308\\
  \end{tabular}
  \caption{Results from Experiment 4. Mean inference time is the mean time spent for inference per period, i.e., the total amount of time spent for 480 periods divided by 480.}
  \label{table:experiment4}
\end{table}

Table \ref{table:experiment4} presents the results for Experiment 4. \emph{Original} refers to the classifier variant used in Experiments 1-3. Column \emph{Inf. time} presents the mean time spent for inference per 60-second period of data in milliseconds (ms). Columns $\kappa$ and \emph{Acc.} present classification performance in terms of the mean $\kappa$ and accuracy across the 10 folds, respectively. Classification performance for the original models is obtained from Experiment 1.2, and therefore shows the mean across the 10 models from 10-fold CV. The float and quantized models are based on the first model from 10-fold CV. Since we only execute models intended for mobile devices on the portable device (float and quantized variants), Table \ref{table:experiment4} does not include inference times for the original models. Instead, we present the size of the original model for comparison. This size depends on the method by which it is stored. We present the two sizes resulting from (1) just saving the weights as a checkpoint, in which case we need to specify the architecture of the model whenever we load it, and (2) saving also the architecture of the model.

Analysis of data from one complete night is in most cases completed in less than a second. This is true even for the float models. We observe almost no reduction in prediction performance of float models compared to original models. For CNNS, the mean performance of the float model is slightly higher than that of the original model. This difference is most likely attributed to the high SD in our results, as seen in Figure \ref{fig:mixed}. The low inference times demonstrate that the classification performance obtained in Experiments 1-3 can be achieved on smartphones. We observe almost no performance reduction for quantized models, even if they reduce inference time by 38.18-51.81\% compared to the float models. This can significantly reduce inference time with larger models or with sensors with higher sampling rates.

\section{Discussion}
The accuracy of classifying Flow data is nearly as high when we train on Nox data as when we train on Flow data. This implies that the need to collect additional data for new, low-quality sensors may not be necessary when high-quality data is available. As a result, the cost of collecting large amounts of sensitive data can be substantially reduced, facilitating low-cost screening based on consumer electronics. Performance improves slightly by training with both low- (Flow) and high-quality (Nox) data, and less high-quality data is required if it is collected simultaneously with the low-quality data.

The simple BLA for Flow data leads to substantial improvements in performance for nearly all ML methods. BLA merely involves standardizing the data in one-minute intervals, showing that significant performance gains can be achieved with only minimal pre-processing. A promising direction for future work is to investigate if additional pre-processing leads to further performance improvements. CNN is the best-performing ML technique, with an accuracy of 0.7609$\pm$0.0284 and nearly the lowest training and testing times. The fact that it performs almost as well with and without BLA distinguishes it from the other ML techniques, and shows that it is by far the most robust ML technique against baseline shifts and wandering. This is not an obvious outcome, since CNN is not designed particularly for time series data like RNN. To the best of our knowledge, our results are the first to demonstrate that this applies also to clinical data from consumer electronics. These results emphasize the versatility of CNN and its usefulness even for physiological time series data. We will use CNN for future work in this area. We obtain similar period-based performance as existing works with low-cost respiratory belts with piezo or bioimpedance technology. For instance, a maximum accuracy of 0.728 and 0.7380 is obtained in \cite{van2020portable, lin2016sleep}. As expected, these results are not at the same level of performance as with PSG and PG with manual scoring, given the much cheaper hardware and the use of generic ML in stead of custom-tailored analysis algorithms. We nevertheless find that automated analysis of Flow data yields predictions that correlate well with actual SA severity, with r-values of up to 0.64, and as such can be used to indicate SA severity. We show that our classifiers can be used on small, resource constrained devices, with almost no loss in classification performance.

The quality of subject-based classification depends on whether we train the classifier with balanced or imbalanced data. Classifiers trained on balanced data most often correctly identify subjects with SA, but tend to over-estimate SA. This is acceptable when false positives can be tolerated to reduce the number of false negatives. Imbalanced training data may nevertheless be preferrable in a home-screening scenario, even if fewer apneic subjects are identified, since it leads to an overall higher prediction accuracy. This reduces the number of unnecessary visits to the doctor and the use of limited medical resources. Imbalanced data also yields the highest increase in the probability of SA upon a positive test result, which is a key indicator of the usefulness of the classifers. This is noteworthy, since applied ML studies commonly use balanced training data. In general, combining balanced and imbalanced training data constitutes a compromise between using balanced and imbalanced data. This approach outperforms the latter two in several ways, e.g., in terms of the r-value (Figure \ref{figure:exp3scatter}), the width of the prediction and confidence bands (Figure \ref{figure:exp3scatter}), and predicting whether a subject has moderate SA (Table \ref{table:binarySC}).

There is possible future work in three areas: (1) We wish to investigate whether our results apply to other types of physiological signals and other data types like images. (2) We wish to investigate alternatives for improving classification performance. Our analysis outlines several promising directions. Improvements in hardware are expected to improve performance, since we use sensors of low quality which impacts performance. Our current data set is relatively small, especially for the low-quality data. We expect the performance to improve with more training data. (3) We wish to investigate personalized training where we leverage data from subjects that are made available at an early stage.

\section{Acknowledgements}
This work was funded by the Norwegian Research Council (as part of the CESAR project, no. 250239/O70), Oslo University Hospital, University of Oslo, and the Norwegian Health Association.

\bibliographystyle{abbrv}
\bibliography{references.bib}

\begin{thebibliography}{10}

\bibitem{website:nox-t3}
Nox t3 sleep monitor, nox medical.
\newblock http://noxmedical.com/products/nox-t3-sleep-monitor, 2020.
\newblock Accessed: 2020-06-16.

\bibitem{al2014classifying}
M.~Al-Mardini, F.~Aloul, A.~Sagahyroon, and L.~Al-Husseini.
\newblock Classifying obstructive sleep apnea using smartphones.
\newblock {\em Journal of biomedical informatics}, 52:251--259, 2014.

\bibitem{alvarez2020machine}
D.~{\'A}lvarez, A.~Cerezo-Hern{\'a}ndez, A.~Crespo, G.~C. Guti{\'e}rrez-Tobal,
  F.~Vaquerizo-Villar, V.~Barroso-Garc{\'\i}a, F.~Moreno, C.~A. Arroyo,
  T.~Ruiz, R.~Hornero, et~al.
\newblock A machine learning-based test for adult sleep apnoea screening at
  home using oximetry and airflow.
\newblock {\em Scientific reports}, 10(1):1--12, 2020.

\bibitem{alvarez2016automated}
D.~{\'A}lvarez, G.~C. Guti{\'e}rrez-Tobal, F.~Vaquerizo-Villar,
  V.~Barroso-Garc{\'\i}a, A.~Crespo, C.~Arroyo, F.~Del~Campo, and R.~Hornero.
\newblock Automated analysis of unattended portable oximetry by means of
  bayesian neural networks to assist in the diagnosis of sleep apnea.
\newblock In {\em 2016 Global Medical Engineering Physics Exchanges/Pan
  American Health Care Exchanges (GMEPE/PAHCE)}, pages 1--4. IEEE, 2016.

\bibitem{alvarez2015computer}
D.~Alvarez-Estevez and V.~Moret-Bonillo.
\newblock Computer-assisted diagnosis of the sleep apnea-hypopnea syndrome: a
  review.
\newblock {\em Sleep disorders}, 2015, 2015.

\bibitem{benjafield2019estimation}
A.~V. Benjafield, N.~T. Ayas, P.~R. Eastwood, R.~Heinzer, M.~S. Ip, M.~J.
  Morrell, C.~M. Nunez, S.~R. Patel, T.~Penzel, J.-L. P{\'e}pin, et~al.
\newblock Estimation of the global prevalence and burden of obstructive sleep
  apnoea: a literature-based analysis.
\newblock {\em The Lancet Respiratory Medicine}, 7(8):687--698, 2019.

\bibitem{berry2012aasm}
R.~B. Berry, R.~Brooks, C.~E. Gamaldo, S.~M. Harding, C.~Marcus, B.~V. Vaughn,
  et~al.
\newblock The aasm manual for the scoring of sleep and associated events.
\newblock {\em Rules, Terminology and Technical Specifications, Darien,
  Illinois, American Academy of Sleep Medicine}, 176:2012, 2012.

\bibitem{camci2017sleep}
B.~Camc{\i}, A.~Y. Kahveci, B.~Arnrich, and C.~Ersoy.
\newblock Sleep apnea detection via smart phones.
\newblock In {\em 2017 25th Signal Processing and Communications Applications
  Conference (SIU)}, pages 1--4. IEEE, 2017.

\bibitem{castillo2019entropy}
Y.~Castillo-Escario, I.~Ferrer-Lluis, J.~M. Montserrat, and R.~Jan{\'e}.
\newblock Entropy analysis of acoustic signals recorded with a smartphone for
  detecting apneas and hypopneas: A comparison with a commercial system for
  home sleep apnea diagnosis.
\newblock {\em IEEE access}, 7:128224--128241, 2019.

\bibitem{chadha1982validation}
T.~Chadha, H.~Watson, S.~Birch, G.~Jenouri, A.~Schneider, M.~Cohn, and
  M.~Sackner.
\newblock Validation of respiratory inductive plethysmography using different
  calibration procedures.
\newblock {\em American Review of Respiratory Disease}, 125(6):644--649, 1982.

\bibitem{chang2019validation}
Y.~Chang, L.~Xu, F.~Han, B.~T. Keenan, E.~Kneeland-Szanto, R.~Zhang, W.~Zhang,
  Y.~Yu, Y.~Zuo, A.~I. Pack, et~al.
\newblock Validation of the nox-t3 portable monitor for diagnosis of
  obstructive sleep apnea in patients with chronic obstructive pulmonary
  disease.
\newblock {\em Journal of Clinical Sleep Medicine}, 15(4):587--596, 2019.

\bibitem{cohen1960}
J.~Cohen.
\newblock A coefficient of agreement for nominal scales.
\newblock {\em Educational and Psychological Measurement}, 20(1):37--46, 1960.

\bibitem{cohn1982respiratory}
M.~Cohn, A.~Rao, M.~Broudy, S.~Birch, H.~Watson, N.~Atkins, B.~Davis, F.~Stott,
  and M.~Sackner.
\newblock The respiratory inductive plethysmograph: a new non-invasive monitor
  of respiration.
\newblock {\em Bulletin europeen de physiopathologie respiratoire}, 18(4):643,
  1982.

\bibitem{collop2011obstructive}
N.~A. Collop, S.~L. Tracy, V.~Kapur, R.~Mehra, D.~Kuhlmann, S.~A. Fleishman,
  and J.~M. Ojile.
\newblock Obstructive sleep apnea devices for out-of-center (ooc) testing:
  technology evaluation.
\newblock {\em Journal of Clinical Sleep Medicine}, 7(5):531--548, 2011.

\bibitem{dehkordi2012monitoring}
P.~Dehkordi, M.~Marzencki, K.~Tavakolian, M.~Kaminska, and B.~Kaminska.
\newblock Monitoring torso acceleration for estimating the respiratory flow and
  efforts for sleep apnea detection.
\newblock In {\em 2012 Annual International Conference of the IEEE Engineering
  in Medicine and Biology Society}, pages 6345--6348. IEEE, 2012.

\bibitem{elmoaqet2020deep}
H.~ElMoaqet, M.~Eid, M.~Glos, M.~Ryalat, and T.~Penzel.
\newblock Deep recurrent neural networks for automatic detection of sleep apnea
  from single channel respiration signals.
\newblock {\em Sensors}, 20(18):5037, 2020.

\bibitem{faust2016review}
O.~Faust, U.~R. Acharya, E.~Ng, and H.~Fujita.
\newblock A review of ecg-based diagnosis support systems for obstructive sleep
  apnea.
\newblock {\em Journal of Mechanics in Medicine and Biology}, 16(01):1640004,
  2016.

\bibitem{ferrer2019automatic}
I.~Ferrer-Lluis, Y.~Castillo-Escario, J.~M. Montserrat, and R.~Jan{\'e}.
\newblock Automatic event detector from smartphone accelerometry: Pilot mhealth
  study for obstructive sleep apnea monitoring at home.
\newblock In {\em 2019 41st Annual International Conference of the IEEE
  Engineering in Medicine and Biology Society (EMBC)}, pages 4990--4993. IEEE,
  2019.

\bibitem{french1999catastrophic}
R.~M. French.
\newblock Catastrophic forgetting in connectionist networks.
\newblock {\em Trends in cognitive sciences}, 3(4):128--135, 1999.

\bibitem{garde2015pulse}
A.~Garde, P.~Dehkordi, D.~Wensley, J.~M. Ansermino, and G.~A. Dumont.
\newblock Pulse oximetry recorded from the phone oximeter for detection of
  obstructive sleep apnea events with and without oxygen desaturation in
  children.
\newblock In {\em 2015 37th Annual International Conference of the IEEE
  Engineering in Medicine and Biology Society (EMBC)}, pages 7692--7695. IEEE,
  2015.

\bibitem{gutierrez2018evaluation}
G.~C. Guti{\'e}rrez-Tobal, D.~{\'A}lvarez, A.~Crespo, F.~del Campo, and
  R.~Hornero.
\newblock Evaluation of machine-learning approaches to estimate sleep apnea
  severity from at-home oximetry recordings.
\newblock {\em IEEE journal of biomedical and health informatics},
  23(2):882--892, 2018.

\bibitem{hafezi2019sleep}
M.~Hafezi, N.~Montazeri, K.~Zhu, H.~Alshaer, A.~Yadollahi, and B.~Taati.
\newblock Sleep apnea severity estimation from respiratory related movements
  using deep learning.
\newblock In {\em 2019 41st Annual International Conference of the IEEE
  Engineering in Medicine and Biology Society (EMBC)}, pages 1601--1604. IEEE,
  2019.

\bibitem{hamborg2020}
M.~Hamborg.
\newblock {Analyzing the Usefulness of a Low-Cost Respiration Sensor for Sleep
  Apnea Detection in a Clinical Setting}.
\newblock Master's thesis, University of Oslo, Department of Informatics, Oslo,
  Norway, 2020.

\bibitem{he2009learning}
H.~He and E.~A. Garcia.
\newblock Learning from imbalanced data.
\newblock {\em IEEE Transactions on knowledge and data engineering},
  21(9):1263--1284, 2009.

\bibitem{hestness2017deep}
J.~Hestness, S.~Narang, N.~Ardalani, G.~Diamos, H.~Jun, H.~Kianinejad,
  M.~Patwary, M.~Ali, Y.~Yang, and Y.~Zhou.
\newblock Deep learning scaling is predictable, empirically.
\newblock {\em arXiv preprint arXiv:1712.00409}, 2017.

\bibitem{hrubos2011norwegian}
H.~Hrubos-str{\o}m, A.~Randby, S.~K. Namtvedt, H.~A. Kristiansen, G.~Einvik,
  J.~Benth, V.~K. Somers, I.~H. Nordhus, M.~B. Russell, T.~Dammen, et~al.
\newblock A norwegian population-based study on the risk and prevalence of
  obstructive sleep apnea the akershus sleep apnea project (asap).
\newblock {\em Journal of sleep research}, 20(1pt2):162--170, 2011.

\bibitem{huang2008clinical}
Q.~Huang, Z.~Qin, S.~Zhang, and C.~Chow.
\newblock Clinical patterns of obstructive sleep apnea and its comorbid
  conditions: a data mining approach.
\newblock {\em Journal of clinical sleep medicine: JCSM: official publication
  of the American Academy of Sleep Medicine}, 4(6):543--550, 2008.

\bibitem{javaid2015towards}
A.~Q. Javaid, C.~M. Noble, R.~Rosenberg, and M.~A. Weitnauer.
\newblock Towards sleep apnea screening with an under-the-mattress ir-uwb radar
  using machine learning.
\newblock In {\em 2015 IEEE 14th International Conference on Machine Learning
  and Applications (ICMLA)}, pages 837--842. IEEE, 2015.

\bibitem{kaimakamis2009screening}
E.~Kaimakamis, C.~Bratsas, L.~Sichletidis, C.~Karvounis, and N.~Maglaveras.
\newblock Screening of patients with obstructive sleep apnea syndrome using c4.
  5 algorithm based on non linear analysis of respiratory signals during sleep.
\newblock In {\em 2009 Annual International Conference of the IEEE Engineering
  in Medicine and Biology Society}, pages 3465--3469. IEEE, 2009.

\bibitem{kristiansen2018data}
S.~Kristiansen, M.~S. Hugaas, V.~Goebel, T.~Plagemann, K.~Nikolaidis, and
  K.~Liest{\o}l.
\newblock Data mining for patient friendly apnea detection.
\newblock {\em IEEE Access}, 6:74598--74615, 2018.

\bibitem{kristiansen2020machine}
S.~Kristiansen, K.~Nikolaidis, T.~Plagemann, V.~Goebel, G.~M. Traaen,
  B.~{\O}verland, L.~Aaker{\o}y, T.~E. Hunt, J.~P. Loennechen, S.~L.
  Steinshamn, C.~H. Bendz, O.-G. Anfinsen, L.~Gullestad, and H.~Akre.
\newblock Machine learning for sleep apnea detection with unattended sleep
  monitoring at home.
\newblock {\em To be published in ACM Transactions on Computing and Healthcare
  (ACM HEALTH)}, 2020.

\bibitem{kristiansen2020comparing}
S.~Kristiansen, G.~M. Traaen, B.~{\O}verland, T.~Plagemann, L.~Gullestad,
  H.~Akre, K.~Nikolaidis, L.~Aaker{\o}y, T.~E. Hunt, J.~P. Loennechen, et~al.
\newblock Comparing manual and automatic scoring of sleep monitoring data from
  portable polygraphy.
\newblock {\em Journal of Sleep Research}, page e13036, 2020.

\bibitem{lin2016sleep}
Y.-Y. Lin, H.-T. Wu, C.-A. Hsu, P.-C. Huang, Y.-H. Huang, and Y.-L. Lo.
\newblock Sleep apnea detection based on thoracic and abdominal movement
  signals of wearable piezoelectric bands.
\newblock {\em IEEE journal of biomedical and health informatics},
  21(6):1533--1545, 2016.

\bibitem{loberg2018quantifying}
F.~L{\o}berg, V.~Goebel, and T.~Plagemann.
\newblock Quantifying the signal quality of low-cost respiratory effort sensors
  for sleep apnea monitoring.
\newblock In {\em Proceedings of the 3rd International Workshop on Multimedia
  for Personal Health and Health Care}, pages 3--11, 2018.

\bibitem{mendoncca2018devices}
F.~Mendon{\c{c}}a, S.~S. Mostafa, A.~G. Ravelo-Garc{\'\i}a, F.~Morgado-Dias,
  and T.~Penzel.
\newblock Devices for home detection of obstructive sleep apnea: A review.
\newblock {\em Sleep medicine reviews}, 41:149--160, 2018.

\bibitem{mendonca2018review}
F.~Mendonca, S.~S. Mostafa, A.~G. Ravelo-Garc{\'\i}a, F.~Morgado-Dias, and
  T.~Penzel.
\newblock A review of obstructive sleep apnea detection approaches.
\newblock {\em IEEE journal of biomedical and health informatics},
  23(2):825--837, 2018.

\bibitem{mostafa2019systematic}
S.~S. Mostafa, F.~Mendon{\c{c}}a, A.~G~Ravelo-Garc{\'\i}a, and F.~Morgado-Dias.
\newblock A systematic review of detecting sleep apnea using deep learning.
\newblock {\em Sensors}, 19(22):4934, 2019.

\bibitem{nandakumar2015contactless}
R.~Nandakumar, S.~Gollakota, and N.~Watson.
\newblock Contactless sleep apnea detection on smartphones.
\newblock In {\em Proceedings of the 13th annual international conference on
  mobile systems, applications, and services}, pages 45--57, 2015.

\bibitem{nepal2002apnea}
K.~Nepal, E.~Biegeleisen, and T.~Ning.
\newblock Apnea detection and respiration rate estimation through parametric
  modelling.
\newblock In {\em Proceedings of the IEEE 28th Annual Northeast Bioengineering
  Conference (IEEE Cat. No. 02CH37342)}, pages 277--278. IEEE, 2002.

\bibitem{pan2009survey}
S.~J. Pan and Q.~Yang.
\newblock A survey on transfer learning.
\newblock {\em IEEE Transactions on knowledge and data engineering},
  22(10):1345--1359, 2009.

\bibitem{pombo2017classification}
N.~Pombo, N.~Garcia, and K.~Bousson.
\newblock Classification techniques on computerized systems to predict and/or
  to detect apnea: A systematic review.
\newblock {\em Computer methods and programs in biomedicine}, 140:265--274,
  2017.

\bibitem{punjabi2008epidemiology}
N.~M. Punjabi.
\newblock The epidemiology of adult obstructive sleep apnea.
\newblock {\em Proceedings of the American Thoracic Society}, 5(2):136--143,
  2008.

\bibitem{sweetzpot2020}
Sweetzpot.
\newblock https://www.sweetzpot.com/.
\newblock Accessed: September 2020.

\bibitem{teran1999association}
J.~Teran-Santos, A.~Jimenez-Gomez, and J.~Cordero-Guevara.
\newblock The association between sleep apnea and the risk of traffic
  accidents.
\newblock {\em New England Journal of Medicine}, 340(11):847--851, 1999.

\bibitem{traaen2020prevalence}
G.~M. Traaen, B.~{\O}verland, L.~Aaker{\o}y, T.~Hunt, C.~Bendz, L.~Sande,
  S.~Aakhus, H.~Zar{\'e}, S.~Steinshamn, O.-G. Anfinsen, et~al.
\newblock Prevalence, risk factors, and type of sleep apnea in patients with
  paroxysmal atrial fibrillation.
\newblock {\em IJC Heart \& Vasculature}, 26:100447, 2020.

\bibitem{tsouti2020development}
V.~Tsouti, A.~Kanaris, K.~Tsoutis, and S.~Chatzandroulis.
\newblock Development of an automated system for obstructive sleep apnea
  treatment based on machine learning and breath effort monitoring.
\newblock {\em Microelectronic Engineering}, page 111376, 2020.

\bibitem{tumer1996estimating}
K.~Tumer and J.~Ghosh.
\newblock Estimating the bayes error rate through classifier combining.
\newblock In {\em Proceedings of 13th International Conference on Pattern
  Recognition}, volume~2, pages 695--699. IEEE, 1996.

\bibitem{uddin2018classification}
M.~Uddin, C.~Chow, and S.~Su.
\newblock Classification methods to detect sleep apnea in adults based on
  respiratory and oximetry signals: a systematic review.
\newblock {\em Physiological measurement}, 39(3):03TR01, 2018.

\bibitem{van2020portable}
T.~Van~Steenkiste, W.~Groenendaal, P.~Dreesen, S.~Lee, S.~Klerkx,
  R.~De~Francisco, D.~Deschrijver, and T.~Dhaene.
\newblock Portable detection of apnea and hypopnea events using bio-impedance
  of the chest and deep learning.
\newblock {\em IEEE Journal of Biomedical and Health Informatics}, 2020.

\bibitem{wang2016attention}
Y.~Wang, M.~Huang, X.~Zhu, and L.~Zhao.
\newblock Attention-based lstm for aspect-level sentiment classification.
\newblock In {\em Proceedings of the 2016 conference on empirical methods in
  natural language processing}, pages 606--615, 2016.

\bibitem{weiss2001effect}
G.~M. Weiss and F.~Provost.
\newblock The effect of class distribution on classifier learning: an empirical
  study.
\newblock 2001.

\bibitem{young2004risk}
T.~Young, J.~Skatrud, and P.~E. Peppard.
\newblock Risk factors for obstructive sleep apnea in adults.
\newblock {\em Jama}, 291(16):2013--2016, 2004.

\bibitem{zhang2013real}
J.~Zhang, Q.~Zhang, Y.~Wang, and C.~Qiu.
\newblock A real-time auto-adjustable smart pillow system for sleep apnea
  detection and treatment.
\newblock In {\em 2013 ACM/IEEE International Conference on Information
  Processing in Sensor Networks (IPSN)}, pages 179--190. IEEE, 2013.

\end{thebibliography}

\end{document}